\pdfoutput=1

\documentclass[11pt]{article}

\usepackage[final]{acl}

\usepackage{times}
\usepackage{latexsym}

\usepackage[T1]{fontenc}

\usepackage[utf8]{inputenc}

\usepackage{microtype}

\usepackage{inconsolata}

\usepackage{graphicx}
\usepackage{booktabs}
\usepackage[most]{tcolorbox}
\usepackage{xcolor}
\usepackage{colortbl}
\usepackage{enumitem}

\title{Dynamic Strategy Planning for Efficient Question Answering with \\ Large Language Models}

\author{Tanmay Parekh$^*$$^\dagger$ \ \ \
    Pradyot Prakash$^\ddagger$ \ \ \
    Alexander Radovic$^\ddagger$ \ \ \
    Akshay Shekher$^\ddagger$ \ \ \
    Denis Savenkov$^\ddagger$ \\
  $^\dagger$ Univeristy of California, Los Angeles \ \ \ \ \
  $^\ddagger$ Meta AI \\
  \texttt{tparekh@cs.ucla.edu}, \texttt{\{pradyot, alexradovic, shekher, denxx\}@meta.com} \\
  }

\begin{document}
\maketitle

\newcommand{\mypar}[1]{\vspace{0.35em}\noindent\textbf{#1}}
\newcommand\blfootnote[1]{%
  \begingroup
  \renewcommand\thefootnote{}\footnote{#1}%
  \addtocounter{footnote}{-1}%
  \endgroup
}

\newcommand{\SideNote}[2]{\todo[color=#1,size=\small]{#2}}
\newcommand{\tanmay}[1]{\SideNote{orange!40}{#1 --tanmay}}

\newcommand{\modelName}[0]{DyPlan}
\newcommand{\modelVerifyName}[0]{DyPlan-verify}

\definecolor{lightyellow}{RGB}{255,235,180}
\definecolor{red}{RGB}{255,130,130}
\definecolor{lightgreen}{RGB}{200,240,200}
\definecolor{mediumgreen}{RGB}{100,230,100}
\definecolor{darkgreen}{RGB}{20,120,20}

\begin{abstract}

Research has shown the effectiveness of reasoning (e.g., Chain-of-Thought), planning (e.g., SelfAsk), and retrieval augmented generation strategies to improve the performance of Large Language Models (LLMs) on various tasks, such as question answering.
However, using a single fixed strategy to answer different kinds of questions is suboptimal in performance and inefficient in terms of generated output tokens and performed retrievals. 
In our work, we propose a novel technique \modelName, to induce a dynamic strategy selection process in LLMs, to improve performance and reduce computational costs in question-answering.
\modelName{} incorporates an initial decision step to select the most suitable strategy conditioned on the input question and guides the LLM's response generation accordingly.
We extend \modelName{} to \modelVerifyName, adding an internal verification and correction process to further enrich the generated answer.
Experiments on three prominent multi-hop question answering (MHQA) datasets reveal how \modelName{} can improve model performance by 7-13\% while reducing the computational cost by 11-32\% relative to the best baseline model.
Code for this work can be found at \url{https://github.com/facebookresearch/dyplan}.
\blfootnote{$^*$Work completed as part of an internship at Meta.}

\end{abstract}

\section{Introduction}

\begin{figure}[t]
    \centering
    \includegraphics[width=0.98\linewidth]{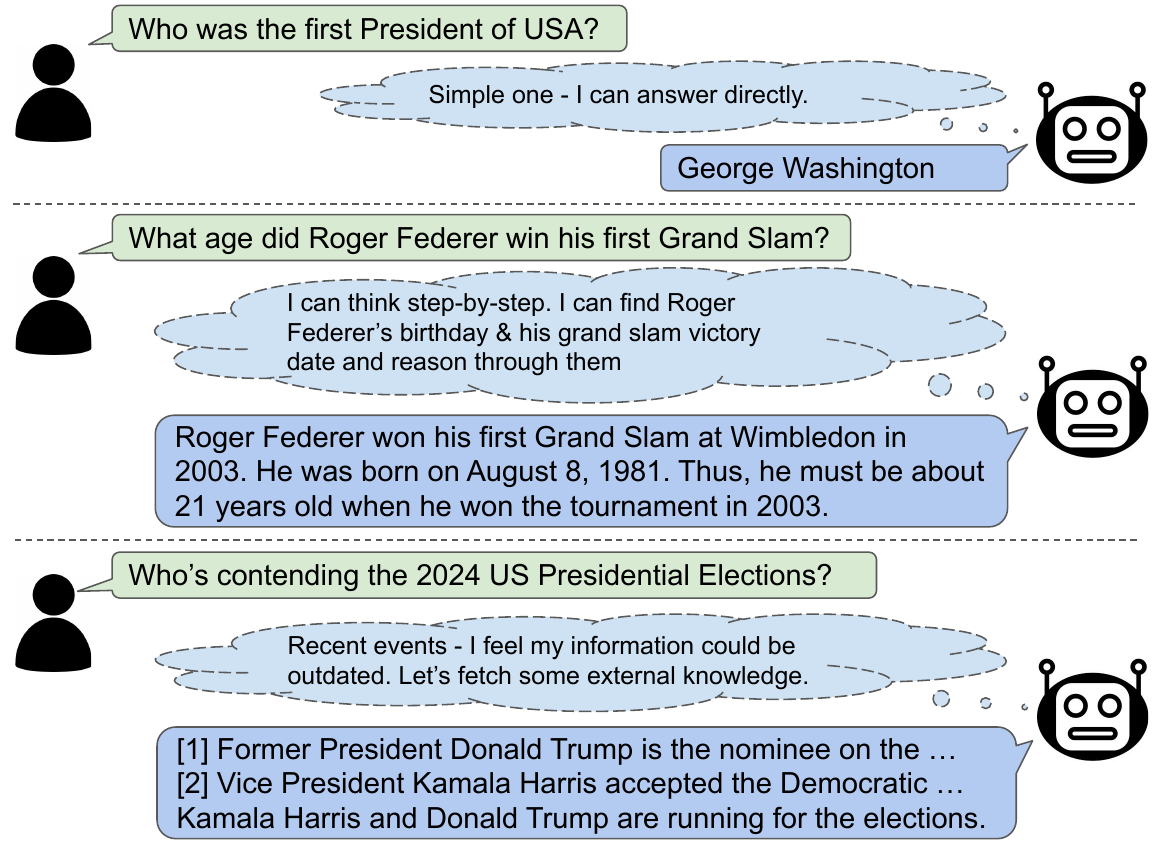}
    \caption{Illustration of dynamically deciding appropriate strategies (indicated by the clouds) conditioned on the input questions.}
    \label{fig:cover-pic}
\end{figure}

Question-answering (QA) for large language models (LLMs) spans a range of question types, from simple queries to those requiring reasoning, external knowledge, step-by-step planning, or a combination of these strategies.
For example (Figure~\ref{fig:cover-pic}), modern LLMs can easily answer \textit{Who was the first president of USA?} but may need some reasoning to figure out \textit{At what age did Roger Federer win his first Grand Slam Title?}, while \textit{Who's contending the 2024 US Presidential Elections?} requires the model to retrieve up-to-date external information.
To this end, previous works have investigated various strategies to induce reasoning, such as Chain of Thought \cite{chain-of-thought}, Tree-of-Thought \cite{tree-of-thought}; or planning, such as SelfAsk \cite{press-etal-2023-measuring}, Decomposed Prompting \cite{decomposed-prompting}, StepBack Prompting \cite{stepback-prompting}; or incorporating external knowledge through Retrieval Augmented Generation \cite{rag} and Knowledge Graphs \cite{kg-in-llms}.

However, employing a single strategy for all different types of questions is sub-optimal as well as quite cost-ineffective in terms of generated tokens and retrievals.
As humans, we rather employ a dynamic decision phase to first determine the most effective strategy before answering the given question.
Similarly, we expect that if the model possesses sufficient self-knowledge to directly answer the question, then `thinking step-by-step' and expending tokens on reasoning is unnecessary.
In other cases, a model may not have enough confidence to answer directly and should spend some computation on additional reasoning.
However, not having enough information about the topic should warrant external retrievals instead.

To this end, we propose to induce a human-like cognitive ability in LLMs through our novel technique, \textbf{\modelName{}} (\textbf{Dy}namic \textbf{Plan}ning).
As illustrated in Figure~\ref{fig:cover-pic}, \modelName{} introduces an initial decision step to select the most suitable strategy conditioned on the input question and then guides the LLM's response generation to use this strategy.
To achieve this behavior, we utilize a multi-turn training paradigm, where the LLM is fine-tuned and calibrated by its own generations.
\modelName{} provides a computationally cost-effective and adaptive solution that efficiently leverages the strengths of various techniques while minimizing computational overhead.

However, there is no complete certainty that the chosen strategy will succeed, as reasoning can be wrong, or retrieved information may turn out to be irrelevant or limited.
At such times, humans usually evaluate and re-select a new strategy to rectify any potential mistakes.
We emulate this internal assessment in LLMs by extending \modelName{} as \textbf{\modelVerifyName{}} (\textbf{Dy}namic \textbf{Plan}ning \& \textbf{Verify}).
Specifically, we add a self-verification step after response generation, which gauges the model's confidence in the provided answer.
If verification fails, the model is prompted to re-select a different strategy, and this cycle can repeat.
Overall, \modelVerifyName{} can achieve higher quality improvements at the cost of additional inference computations.

To evaluate the efficacy of our proposed techniques, we benchmark them on three QA datasets - HotpotQA \cite{yang-etal-2018-hotpotqa}, 2WikiMultihopQA \cite{ho-etal-2020-constructing}, and Musique \cite{trivedi-etal-2022-musique-updated}.
We consider four primary strategies: (1) \textit{Direct} answering directly, (2) \textit{Reason} utilizing Chain-of-Thought \cite{chain-of-thought}, (3) \textit{Plan} leveraging SelfAsk \cite{press-etal-2023-measuring}, and (4) \textit{Retrieval} using external knowledge with RAG \cite{rag}.
We use the LLaMa3-8B model \cite{llama3} as our base model.
We majorly compare against fine-tuned LLMs utilizing fixed strategies along with other ensemble and dynamic thinking baselines.
Results reveal that \modelName{} reduces computational costs by 26-32\% along with performance gains of 7\% averaged across the datasets over the best baseline.
\modelVerifyName{} further improves performance to an average of 12-13\% while providing 11-19\% computational cost reductions.
Analyses reveal how \modelName{} is better calibrated and generalizable and provide insights into its decision-making and verification ability.

In conclusion, we make these contributions:
(1) we propose dynamic strategy planning through \modelName{} to improve the performance and computational cost-efficiency of LLMs for QA,
(2) we extend \modelName{} to \modelVerifyName{} introducing verification of correctness to further boost model performance,
(3) we conduct extensive experiments and analyses using four major strategies on three complex QA datasets to demonstrate \modelName's cost-effectiveness and strong performance.

\begin{figure}[t]
    \centering
    \includegraphics[width=0.8\linewidth]{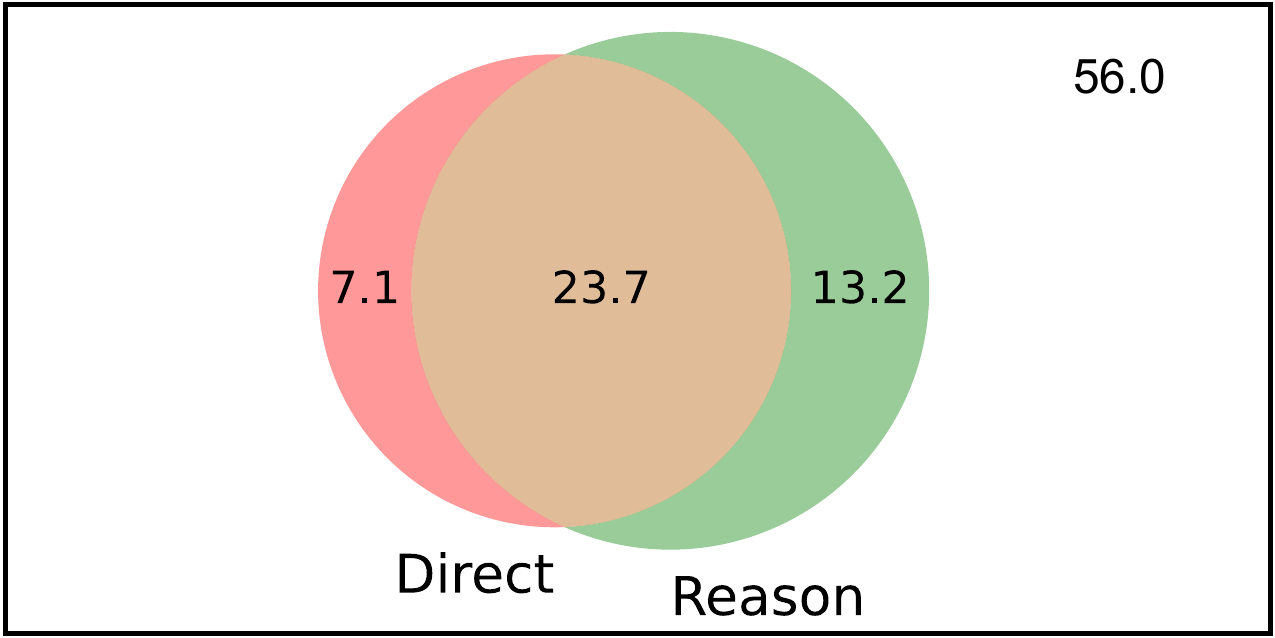}
    \caption{Venn Diagram representing the F1 contribution of Direct and Reason strategies for HotpotQA.}
    \label{fig:2wiki-direct-reason-venn}
\end{figure}

\section{Methodology}
\label{sec:methodology}

To mimic cost-effective human cognitive thinking in LLMs, we propose our novel technique - \textbf{\modelName{}}.
Unlike traditional approaches that rely on a single fixed strategy for all questions, our technique employs dynamic strategy planning to determine the most effective approach for each question.
We extend \modelName{} to \textbf{\modelVerifyName{}} by incorporating additional verification and re-attempting the question with alternative strategies if necessary.
We first motivate the potential impact of dynamic strategy planning in \S~\ref{sec:methodology-motivation} and later provide specific details about our techniques.

\begin{figure*}[t]
    \centering
    \includegraphics[width=\linewidth]{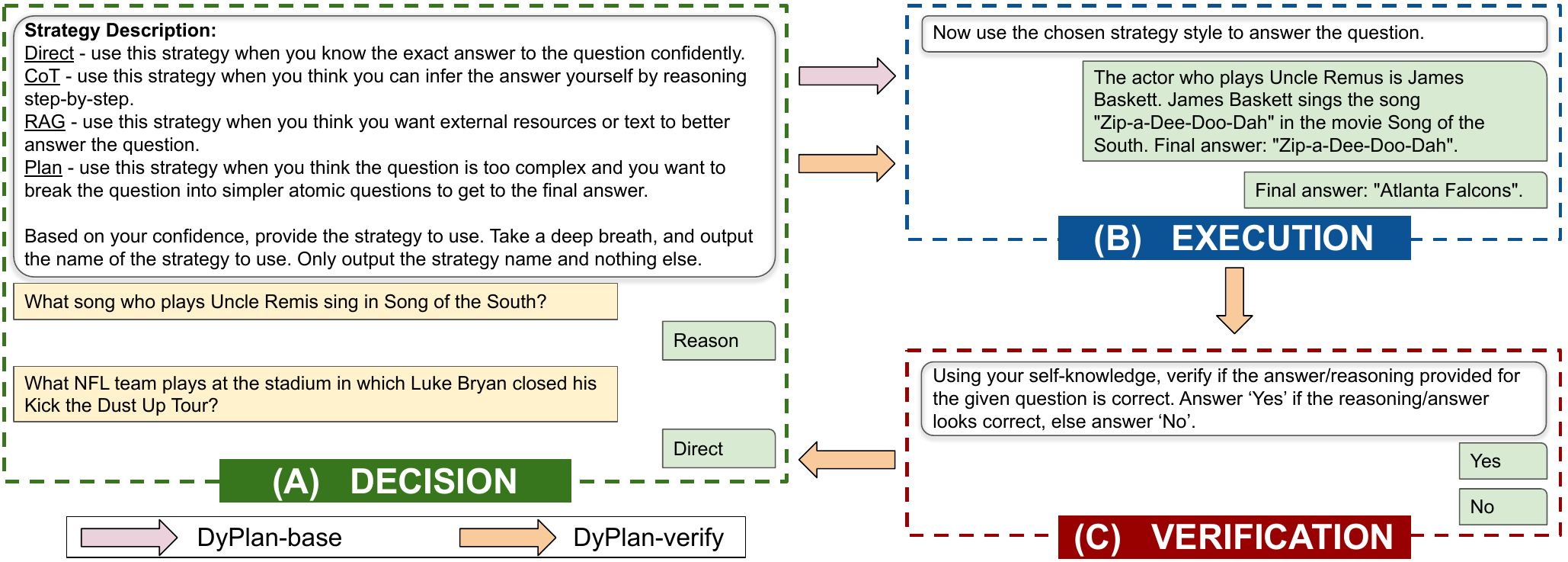}
    \caption{The different components in \modelName{} along with the pipeline flow for two example questions. The Decision component chooses an appropriate strategy from a pool of strategies. The Execution component runs the chosen strategy. The Verification component (used in \modelVerifyName{}) self-verifies the correctness of the provided answer.}
    \label{fig:methodology}
\end{figure*}

\subsection{Motivation}
\label{sec:methodology-motivation}

Dynamic strategy planning can help reduce inference computational costs by simply selecting lower-cost strategies to answer simpler questions.
In our work, we additionally posit that it can improve model performance as strategy selection can act as an ensembling method.
We verify this hypothesis through a simple analysis using two strategies of \textit{Direct} and \textit{Reason}.
Direct prompts the LLM to directly provide the answer to the question while Reason utilizes Chain-of-Thought \cite{chain-of-thought} to answer step-by-step.
Utilizing these strategies, we prompt Llama3-8B-Instruct \cite{llama3}, evaluate using F1 score on 1000 samples from 2WikiMultihopQA \cite{ho-etal-2020-constructing} dataset, and show their performances as a Venn diagram in Figure~\ref{fig:2wiki-direct-reason-venn}.
Generally, one can assume that adding reasoning should only allow us to further answer harder questions while still being able to answer all questions that Direct strategy answers correctly.
However, there is a significant contribution of 7.1\% F1 where Reason is incorrect, but Direct is correct (for example, a mistake or hallucination can lead to an incorrect result).
We provide a qualitative example in Table~\ref{tab:motivation-example} highlighting this behavior (more examples in Table~\ref{tab:strategy-violation-examples}).
We notice similar patterns across datasets and strategies (shown in \S~\ref{sec:appendix-hierarchy-violations}).
Such patterns shed light on how the choice of an appropriate strategy can also improve model performance.

\begin{table}[t]
    \centering
    \small
    \begin{tabular}{p{7.2cm}}
        \toprule
        \textbf{Question:} Which band, Letters to Cleo or Screaming Trees, had more members? \\
        \textbf{Gold Answer:} Letters to Cleo \\
        \midrule
        \textbf{Direct:} Final answer: "Letters to Cleo" \\
        \midrule
        \textbf{Reason:} Letters to Cleo had 4 members: Brian Desveaux, Michael Hubbard, Ken Ray, and Stacy Jones. Screaming Trees had 4 members: Mark Lanegan, Gary Lee Conner, Van Conner, and Barrett Martin. Both bands had the same number of members, 4. Final answer: "same" \\
        \bottomrule
    \end{tabular}
    \caption{Qualitative example highlighting how LLMs can yield the right answer by directly answering the question, while reasoning can lead to incorrect answers.}
    \label{tab:motivation-example}
\end{table}


\subsection{\modelName{} Components}

Our techniques majorly utilize three components in a plug-and-play manner: (1) Decision, which selects a strategy to follow; (2) Execution, which generates the answer using the chosen strategy; and (3) Verification, which evaluates the answer's correctness.
We describe them in detail below and provide a high-level overview diagram in Figure~\ref{fig:methodology}.

\paragraph{Decision:}
The Decision component is the core of our technique, responsible for dynamically selecting the optimal strategy for a given question.
This is achieved by presenting an LLM with a strategy pool with their descriptions and prompting it to leverage its self-confidence to choose the most suitable and efficient strategy.
This component provides an opportunity to optimize efficiency while still enabling powerful reasoning to improve performance when possible, unlike OpenAI's o1 model,\footnote{\url{https://platform.openai.com/docs/models/o1}} which currently applies thinking even for simple questions.
We provide an illustration prompt of this component in Figure~\ref{fig:methodology}(A).


\paragraph{Execution:}
The Execution component involves prompting the model to apply the selected strategy from the Decision component to generate an answer to the question.
Although analogous to fixed-strategy prompting, our approach is different since we enable dynamic strategy execution based on the previously chosen strategy.
We provide an illustration Execution prompt in Figure~\ref{fig:methodology}(B).


\paragraph{Verification:}
The Verification component is optional and only part of our extended technique \modelVerifyName{}.
Intuitively, when we make a decision to choose a certain strategy, like reasoning or retrieval, we cannot be certain it will be successful, as reasoning may fail and retrieval may get irrelevant results.
Therefore, \modelName{} should have the ability to correct the course as long as we have some more budget before having to present the final answer.
This component assesses the validity of the Execution output by prompting the LLM to leverage its self-knowledge and confidence to evaluate the answer's reasonableness and correctness.
To minimize computational cost, we implement this component by asking the LLM to simply output yes/no, as shown in Figure~\ref{fig:methodology}(C).


\subsection{\modelName{} Pipeline Flow}
\label{sec:pipeline-flow}

\modelName{} majorly achieves computational cost minimization by dynamic decision-making in the Decision phase and restricting generation output space for each component.
The base version of \modelName{} (also referred to as \modelName-base) employs a low-cost Decision-Execution pipeline (pink arrow in Figure~\ref{fig:methodology}).
On the other hand, our extension technique \modelVerifyName{} utilizes an iterative loop of Decision-Execution-Verification (orange arrow in Figure~\ref{fig:methodology}).
If verification fails, the pipeline reverts to the Decision component to select an alternative strategy; otherwise, it exits the loop with the execution answer.
This iterative loop runs for a preset number of rounds based on the inference budget or until the Decision runs out of usable strategies.
Both pipelines are implemented using multi-turn chat, with each component corresponding to a single turn.

\section{Data Creation for Finetuning}
\label{sec:data-creation}


To adhere LLMs with the \modelName{} pipeline, we fine-tune the LLM on \modelName-specific data.
To ensure zero human annotation cost, we propose automatic data creation utilizing an existing QA dataset $\mathcal{D}$ and a strategy set $\mathcal{S}$ comprising $n$ strategies.
We order $\mathcal{S}= [s_1, \dots, s_n]$ by strategy preference, with $s_1$ being the most preferred and $s_n$ the least preferred (e.g. in order of computational cost-efficiency/performance).
For each strategy $s\in\mathcal{S}$, we prompt the base LLM on all datapoints $d\in\mathcal{D}$ and evaluate the results against the ground truth.
This yields two disjoint subsets for each $s$: $D^s_p$, comprising datapoints where $s$ produced the correct answer, and $D^s_n$, comprising the remaining datapoints where $s$ failed to produce the correct answer.
Utilizing the base LLM (instead of distilling from larger LLMs) ensures a stronger model self-calibration.
Using the positive and negative subsets, we create component-specific data for \modelName{} (described below) and train the LLM on the combination of all the component data.
We conduct training only on the last-turn response for the multi-turn training instances.

\paragraph{Decision:}
\label{sec:methodology-decision}
We define an optimal mapping function $f^* : \mathcal{D} \rightarrow \mathcal{S}$ that assigns each training datapoint $d \in \mathcal{D}$ to the first strategy $s \in \mathcal{S}$ (according to the preference order) that yields the correct answer.
If none of the strategies produce the correct answer, $d$ is mapped to the least preferred strategy $s_n$.
The Decision component's training data consists of mapped input-output pairs $(d, f^*(d))$.

\paragraph{Execution:}
The input here is a multi-turn chat where the first turn (Decision) selects a strategy $s$.
In the second turn (Execution), the output is set as the base LLM generation using strategy $s$.
Utilizing the base LLM response aids efficient and faster model training.
To minimize noise, we utilize only the positive data $D^s_p$ for each strategy $s$.

\paragraph{Verification:}
For this component, we create binary training data by mapping positive data $D^s_p$ to "yes" and negative data $D^s_n$ to "no".
Multi-turn traces are generated by forcing the selection of strategy $s$ in the first turn (Decision) and using the base LLM response in the second turn (Execution).

\begin{figure}
    \centering
    \includegraphics[width=\linewidth]{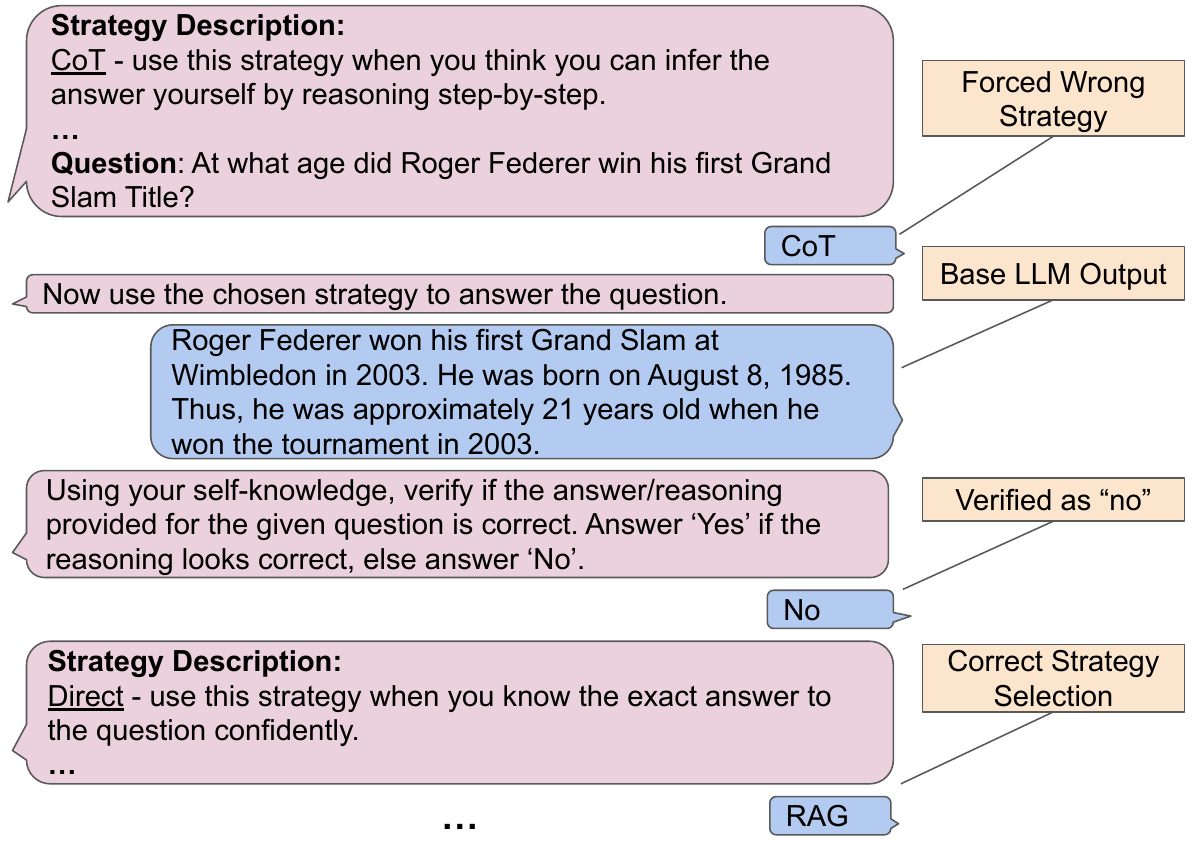}
    \caption{Illustration of an automatically created multi-turn data instance utilizing a forced wrong strategy in the first round with the correct strategy in the second round. LLM is only trained on the second round.}
    \label{fig:multi-turn-data-creation}
\end{figure}

\paragraph{Multi-round data:}
To facilitate multiple rounds of the Decision-Execution-Verification pipeline for \modelVerifyName, we generate additional training data for each component using a reduced dataset $\mathcal{D}' \subset \mathcal{D}$.
Specifically, $\mathcal{D}'$ comprises subsets $\mathcal{D}^{s_i, s_j}_{n,p}$, where each datapoint $d \in \mathcal{D}^{s_i, s_j}_{n,p}$ satisfies $d \in \mathcal{D}^{s_i}_n$ and $d \in \mathcal{D}^{s_j}_p$.
In other words, strategy $s_i$ is incorrect for $d$ and is used as the wrong strategy in the first round, while the correct strategy $s_j$ is used in the second round.
We provide an illustration of such a two-round training instance in Figure~\ref{fig:multi-turn-data-creation}.

\section{Experimentation Details}

We describe the benchmarking datasets and the evaluation metrics. Next, we discuss the strategies, baselines, and, finally, the implementation details.

\paragraph{Benchmarking Datasets:}
LLMs seem to perform well on simpler question-answering datasets like SQuAD \cite{mhqa-survey}.
Instead, we consider three complex Wikipedia-based multi-hop QA (MHQA) datasets to benchmark the performance of our technique, namely HotpotQA \cite{yang-etal-2018-hotpotqa}, 2WikiMultihopQA (2WikiQA) \cite{ho-etal-2020-constructing}, and Musique \cite{trivedi-etal-2022-musique-updated}.
HotpotQA was one of the first human-created MHQA datasets with upto 2-hop reasoning questions.
2WikiMultihopQA further improved over HotpotQA by improving the complexity and reasoning depth of the questions.
Musique is a rule-based constructed dataset created by composing different single-hop questions.
These unique challenges of each dataset aid extensive benchmarking.
We utilize 1000 samples from the development sets of these datasets as the main evaluation dataset.

\paragraph{Evaluation Metrics:}
We evaluate the models on two major dimensions of \textit{performance} and \textit{computational cost}.
For performance, we utilize \textbf{Exact Match (EM)} and \textbf{F1 score} evaluated against the ground truth - higher the better.
For computational cost, we consider the \textbf{number of generated tokens (\# T)} and \textbf{number of retrievals (\# R)} - lower the better.
For \modelName, we report the aggregated cost metrics across the turns.

\begin{table*}[t]
    \centering
    \small
    \begin{tabular}{p{3cm}|cccc|cccc|cccc}
        \toprule
        \textbf{Technique}  & \multicolumn{4}{c|}{\textbf{HotpotQA}} & \multicolumn{4}{c|}{\textbf{2WikiMultihopQA}} & \multicolumn{4}{c}{\textbf{Musique}} \\
        & \textbf{EM} & \textbf{F1} & \textbf{\# T} & \textbf{\# R} & \textbf{EM} & \textbf{F1} & \textbf{\# T} & \textbf{\# R} & \textbf{EM} & \textbf{F1} & \textbf{\# T} & \textbf{\# R} \\
        \midrule
        Fixed-base Direct  & 23.8   & 32.3   & 95    & 0  & 32.1     & 37.4     & 65     & 0    & 2.3   & 9.3    & 99    & 0  \\
        Fixed-base Reason   & 27.2   & 37.5   & 124   & 0  & 19.7     & 27.4     & 65     & 0    & 7.2   & 16.7   & 129   & 0  \\
        Fixed-base Plan   & 24.1   & 33.8   & 203   & 0  & 25.4     & 31.5     & 197    & 0    & 5.8   & 13.4   & 203   & 0  \\
        Fixed-base Retrieval & 36.1   & 47.9   & 185   & 1  & 31.6     & 40.4     & 101    & 1    & 9.6   & 18     & 187   & 1 \\
        \midrule
        Fixed-sft Direct  & 24.1   & 34.3   & 9    & 0  & 32.6     & 38.4     & 10     & 0    & 2.4   & 9    & 17    & 0  \\
        Fixed-sft Reason  & 27.6   & 37.9   & 53   & 0  & 29.3     & 35.6     & 77     & 0    & 7.6   & 16.4   & 63   & 0  \\
        Fixed-sft Plan  & 26.3   & 36   & 105   & 0  & 26.9     & 34.7     & 116    & 0    & 6.6   & 15   & 117   & 0  \\
        Fixed-sft Retrieval [ref] & \underline{36.8}   & \underline{48.6}   & 53   & 1  & 32.8     & 40.0     & 56    & 1    & 9.3   & 18.4     & 88   & 1 \\
        \midrule
        Classifier & 32.6 & 43.9 & \textbf{34} & \textbf{0.59} & 36.0 & 43.1 & \textbf{28} & \textbf{0.45} & 8.0 & 17.5 & 82 & \textbf{0.90} \\
        Ensemble & 35.9 & 47.5 & 220 & 1 & 35.7 & 42.8 & 260 & 1 & 8.8 & 18.1 & 279 & 1 \\
        \textbf{\modelName-base (ours)} & 36.1 & 47.6 & \underline{42} & \underline{0.76} & \underline{37.8} & \underline{46.0} & \textbf{28} & \underline{0.48} & 10.1 & 19.8  & \textbf{65} & \underline{0.98} \\
        \textbf{\modelVerifyName{} (ours)} & 36.7 & 48.5 & 53 & 0.79 & \textbf{40.5} & \textbf{49.6} & \underline{45} & 0.65 & \underline{10.8} & \underline{20.4} & \underline{77} & 0.99 \\
        \midrule
        ReAct & 20.5 & 27.5 & 255 & 3.91 & 27.9 & 32.3 & 226 & 3.01 & 4.4 & 8.5 & 290 & 5.10 \\
        DRAGIN & \textbf{38.9} & \textbf{50.2} & 724 & 2.23 & 32.7 & 41.8 & 272 & 1.67 & \textbf{11.9} & \textbf{22.0} & 993 & 3.03 \\
        \bottomrule
    \end{tabular}
    \caption{The main results comparing \modelName{} and \modelVerifyName{} with other baselines. We mark the best and second-best metrics in \textbf{bold} and \underline{underline}. [ref] indicates the main reference baseline.}
    \label{tab:main-results}
\end{table*}

\paragraph{Strategies:}
We focus on four major themes of strategies, as follows:
\begin{enumerate}[itemsep=0pt]
    \item \textit{Direct}: LLM is prompted to directly provide the final answer. This is the cheapest strategy in terms of computational cost.
    \item \textit{Reason}: LLM is prompted to reason to reach the final answer. We utilize Chain-of-Thought (CoT) \cite{chain-of-thought} to reason step-by-step. This strategy is more expensive than \textit{Direct} in terms of generated tokens.
    \item \textit{Plan}: LLM is prompted to decompose the question as part of planning and reason through the atomic questions to reach the final answer. We utilize SelfAsk \cite{press-etal-2023-measuring} as a prototype for this strategy. This is the most expensive in terms of generated tokens.
    \item \textit{Retrieval}: Following RAG \cite{rag}, using the question as the query, three external passages retrieved from Wikipedia are fed to the LLM. LLM is prompted to reason to reach the final answer. This strategy is expensive in terms of retrievals.
\end{enumerate}

\paragraph{Baseline Models:}
As baselines, we consider:
(1) \textit{Fixed-base} prompts the base LLM with a single fixed strategy,
(2) \textit{Fixed-sft} prompts a LLM fine-tuned on the fixed strategy using the positive base LLM traces,
(3) \textit{Classifier} trains an external classifier to choose the strategy and chooses the corresponding fine-tuned LLM response,
(4) \textit{Ensemble} simply outputs the majority ensemble using all the Fixed-sft strategy responses.

Additionally, we consider some similar works utilizing dynamic decision-making as reference such as:
(5) \textit{ReAct} \cite{react} uses thoughts-actions-observation tuples to guide model generation.
(6) \textit{DRAGIN} \cite{su-etal-2024-dragin} utilizes dynamic retrieval based on model entropy.
Both these baselines are orthogonal to our work and can be utilized in a complementary manner as individual strategies for \modelName.
We majorly compare the cost-effectiveness of \modelName{} with these techniques.

\paragraph{Implementation Details:}
For all experiments, we utilize the LLaMa3-8B-Instruct model \cite{llama3} as the base LLM.
We set the strategy order in increasing order of model performance as Direct-Plan-Reason-Retrieval for training data creation.
We use Low-Rank Adaptation \cite{lora} with rank 32 using LLaMa-Factory \cite{zheng-etal-2024-llamafactory} for fine-tuning the base LLM. 
We utilize code from DRAGIN \cite{su-etal-2024-dragin} to implement the fixed strategy baselines as well as evaluate our techniques.
Our reported numbers are averaged scores over three runs.
Additional details and hyperparameters are provided in Appendix~\ref{sec:hyperparams}.

\begin{figure}[t]
    \centering
    \includegraphics[width=\linewidth]{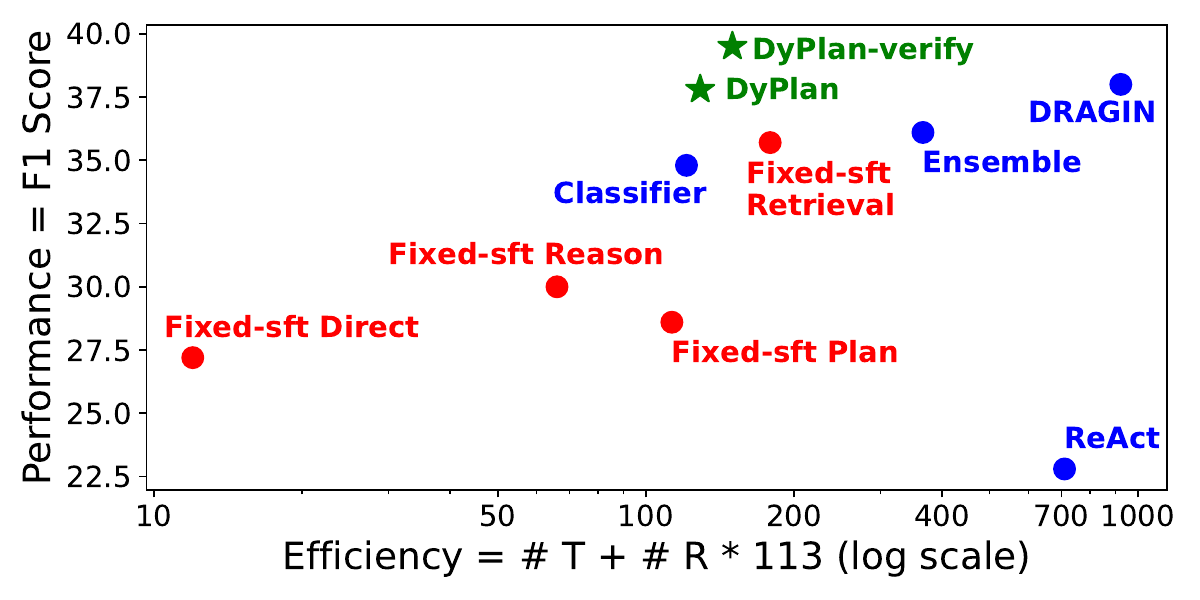}
    \caption{Performance v/s Inference Efficiency for various techniques. Our technique \modelName{} (\textcolor{darkgreen}{in green}) provides the best performance while also reducing the inference costs relative to Fixed-sft Retrieval baseline.}
    \label{fig:main-avg-plot}
\end{figure}

\begin{table*}[t]
    \centering
    \small
    \begin{tabular}{p{3.6cm}|cccc|cccc|cccc}
        \toprule
        \textbf{Technique}  & \multicolumn{4}{c|}{\textbf{HotpotQA}} & \multicolumn{4}{c|}{\textbf{2WikiMultihopQA}} & \multicolumn{4}{c}{\textbf{Musique}} \\
        & \textbf{EM} & \textbf{F1} & \textbf{\# T} & \textbf{\# R} & \textbf{EM} & \textbf{F1} & \textbf{\# T} & \textbf{\# R} & \textbf{EM} & \textbf{F1} & \textbf{\# T} & \textbf{\# R} \\
        \midrule
        Fixed-sft Direct  & 24.1   & 34.3   & 9    & -  & 32.6     & 38.4     & 10     & -    & 2.4   & 9    & 17    & -  \\
        Fixed-sft Reason [ref1]  & 27.6   & 37.9   & 53   & -  & 29.3     & 35.6     & 77     & -    & 7.6   & 16.4   & 63   & -  \\
        Fixed-sft Plan  & 26.3   & 36   & 105   & -  & 26.9     & 34.7     & 116    & -    & 6.6   & 15   & 117   & -  \\
        Fixed-sft Retrieval [ref2] & 36.8   & 48.6   & 53   & 1  & 32.8     & 40.0     & 56    & 1    & 9.3   & 18.4     & 88   & 1 \\
        \midrule
        \multicolumn{13}{c}{\textbf{Strategy Combination: Direct - Plan - Reason} \quad \quad \quad \quad \textbf{Reference: Fixed-sft Reason}} \\
        \midrule
        Classifier & 26.3 & 36.4 & 73 & - & 31.7 & 38.1 & 57 & - & 6.3 & 14.6 & 115 & - \\
        Ensemble & 27.6 & 37.9 & 167 & - & 29.3 & 35.6 & 203 & - & 7.6 & 16.4 & 191 & - \\
        \textbf{\modelName-base (ours)} & 28.0 & 38.2 & \textbf{47} & - & 33.5 & 41.3 & \textbf{36} & - & \textbf{8.1} & 16.7 & \textbf{67} & - \\
        \textbf{\modelVerifyName{} (ours)} & \textbf{28.3} & \textbf{38.8} & 59 & - & \textbf{37.4} & \textbf{43.8} & 68 & - & 7.9 & \textbf{16.8} & 164 & - \\
        \midrule
        \multicolumn{13}{c}{\textbf{Strategy Combination: Reason - Retrieval} \quad \quad \quad \quad \textbf{Reference: Fixed-sft Retrieval}} \\
        \midrule
        Classifier & 32.7 & 44.3 & 53 & \textbf{0.52} & 31.4 & 38.2 & 57 & \textbf{0.54} & 9.1 & 18.2 & 88 & 0.98 \\
        Ensemble & 36.8 & 48.6 & 106 & 1 & 32.8 & 40.0 & 134 & 1 & 9.3 & 18.4 & 151 & 1 \\
        \textbf{\modelName-base (ours)} & 35.5 & 47.3 & \textbf{51} & 0.71 & 34.5 & 43.8 & \textbf{50} & 0.6 & \textbf{10.8} & 20.5 & \textbf{65} & \textbf{0.97} \\
        \textbf{\modelVerifyName{} (ours)} & \textbf{37.2} & \textbf{49.4} & 92 & 0.79 & \textbf{37.4} & \textbf{46.0} & 57 & 0.74 & 10.6 & \textbf{20.6} & 71 & \textbf{0.97} \\
        \bottomrule
    \end{tabular}
    \caption{Performance and computational cost metrics for two strategy combinations of Direct-Plan-Reason ([ref1] is reference) and Reason-Retrieval ([ref2] is reference). We mark the best and second-best metrics in \textbf{bold} and \underline{underline}.}
    \label{tab:other-strategy-combinations}
\end{table*}

\section{Results}

We present our main results comparing \modelName{} utilizing all the strategies with other baselines in Table~\ref{tab:main-results}.
We utilize the best-performing Fixed-sft Retrieval model as the reference baseline for comparisons.
We also aggregate these metrics across datasets and plot Performance (F1 score) v/s Efficiency (weighted sum of \# T and \# R)\footnote{Weights are determined based on pricing of input and output tokens for GPT4o-mini.} in Figure~\ref{fig:main-avg-plot}.

We note that external classifiers help reduce computational cost but don't improve model performance - demonstrating the difficulty of the task.
Dynamic decision-making frameworks like DRAGIN and Ensemble improve performance but are 2-5x more expensive.
To this end, \modelName{} provides the best balance with an average reduction of  \textbf{32\% token and 26\% retrieval cost} along with relative performance improvements of \textbf{7\% EM and 7\% F1}.
\modelVerifyName{} further improves performance with average relative gains of \textbf{13\% EM and 12\% F1} while reducing the token and retrieval cost by \textbf{11\% and 19\%} respectively.
In the best case scenario on 2WikiMultihopQA, \modelName{} shows 16\% performance gains with 52\% computational cost reductions, and \modelVerifyName{} shows 24\% performance gains while reducing costs by 35\%.
Overall, \modelName{} provides strong computational cost reductions along with decent performance gains.


\subsection{Other strategy combinations}

To demonstrate the generalizability of our technique across strategy combinations, we consider two additional combinations of strategies.
The first combination - Direct, Plan, Reason - explores the ability of LLMs to utilize only their self-knowledge to answer the question.
The second combination - Reason and Retrieval - explores the LLM's calibration to decide if it requires any external information to answer the question.
We show the results for these combinations in Table~\ref{tab:other-strategy-combinations}.
Similar to the main results, we observe the superior performance of \modelName{} and \modelVerifyName{} in terms of model performance (average relative gains of 6\%-10\%) as well as computational cost reduction (average reduction of 13\%-20\% tokens and 17\%-24\% retrievals).

\section{Analysis}

We conduct additional experiments to better understand the quality of \modelName{} decision-making and verification and its generalization across datasets.

\subsection{Calibration Analysis}
\label{sec:calibration-analysis}

In \S~\ref{sec:methodology-decision}, we defined an optimal policy $f^*$ for each question as a strategy to pick the most cost-effective technique that yields the correct answer. 
Here, we analyze model calibration, that is, how well its decisions align with the optimal policy at test time (additional details are provided in \S~\ref{sec:appendix-expts}). 


\begin{figure}
    \centering
    \includegraphics[width=\linewidth]{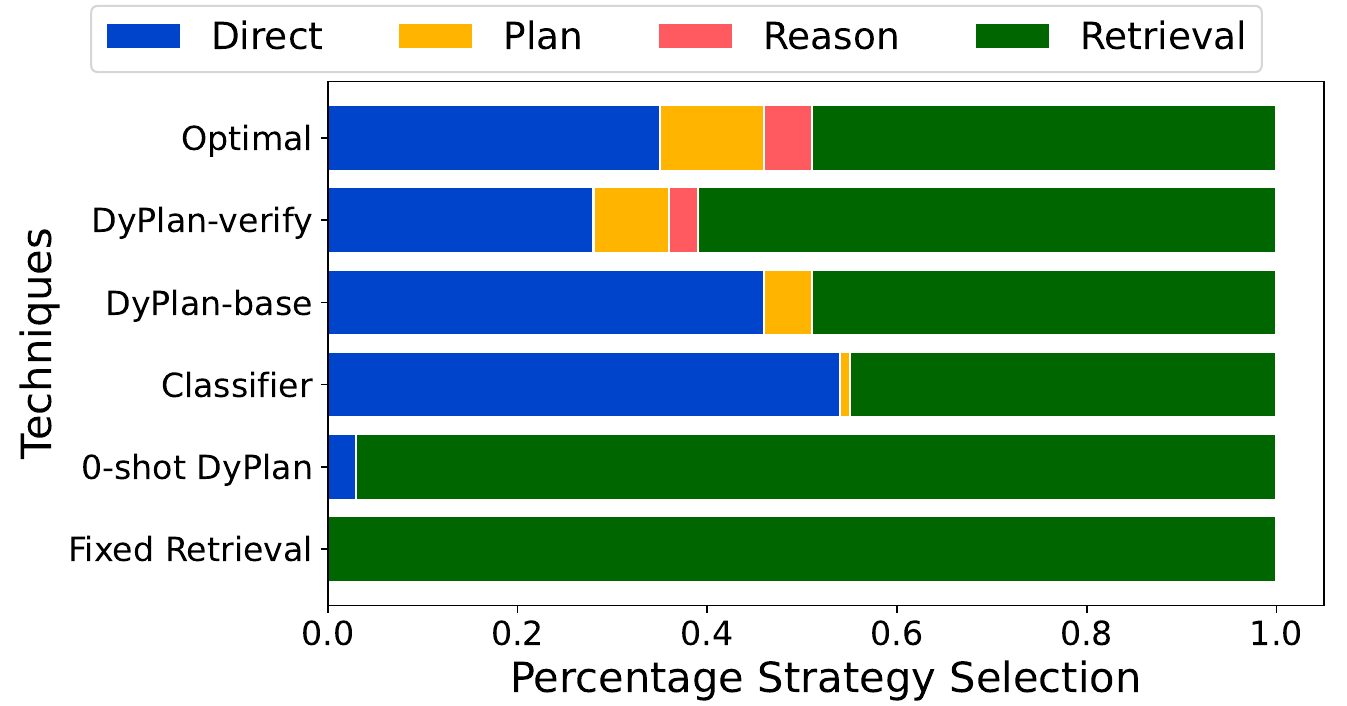}
    \caption{Comparing the strategy planning distribution of various techniques with optimal policy for 2WikiQA.}
    \label{fig:decision-analysis}
\end{figure}

\subsubsection{Decision component of \modelName}
\label{sec:decision-making-analysis}

We compare the strategy planning distribution of various techniques with the optimal policy for 2WikiMultihopQA in Figure~\ref{fig:decision-analysis}.
The major difference is the usage of Plan and Reason which are nearly 0\% for Fixed/Classifier approaches.
On the other hand, \modelName-base and \modelVerifyName{} are closer to the optimal distribution.
We quantify this proximity of the probability distributions in terms of KL divergence.
\modelName-base and \modelVerifyName{} achieve low scores of 0.066 and 0.014, respectively, while the classifier baseline has a high divergence score of 0.35.
Finally, for a stronger sanity check, we compute the accuracies of the strategy choice (relative to optimal policy) of \modelName{} in Table~\ref{tab:decision-accuracies}.
The high improvements relative to other baselines highlight the better strategy planning and stronger calibration of \modelName{}, while throwing light towards further possible improvements.

\begin{table}[t]
    \centering
    \small
    \begin{tabular}{lc}
        \toprule
        \textbf{Technique} & \textbf{Accuracy} \\
        \midrule
        Random & 25.8\% \\
        Majority & 30.8\% \\
        Classifier & 48.1\% \\
        \modelName & 61.7\% \\
        \bottomrule
    \end{tabular}
    \caption{Accuracy of the Decision component of \modelName{} in choosing the right strategy evaluated using the optimal policy on the HotpotQA dataset.}
    \label{tab:decision-accuracies}
\end{table}

\begin{table}[t]
    \centering
    \small
    \setlength\tabcolsep{4pt}
    \begin{tabular}{l|cccc}
        \toprule
        \textbf{Dataset} & \textbf{KL-pre} & \textbf{KL-post} & \textbf{Reject \%} & \textbf{Ver Prec} \\
        \midrule
        HotpotQA & 0.281 & 0.068 & 8\% & 80\% \\
        2WikiQA & 0.138 & 0.014 & 13\% & 71\% \\
        Musique & 0.240 & 0.001 & 16\% & 97\% \\
        \bottomrule
    \end{tabular}
    \caption{Studying the impact of verification in \modelVerifyName{} by evaluating the strategy usage KL divergence with the optimal policy pre (KL-pre) and post (KL-post) verification, the \% datapoints verified as ``no" (Reject \%) and the verification precision of ``no" (Ver Prec).}
    \label{tab:verification-stats}
\end{table}

 \subsubsection{Verification analysis of \modelVerifyName}

We study the verification precision, answer rejection rate (\% datapoints verified as ``no"), and the change in strategy distribution pre and post-verification (in terms of KL divergence relative to optimal strategy) to gain a deeper understanding of the impact of verification in \modelVerifyName.
We provide these statistics in Table~\ref{tab:verification-stats}.
The huge drops in KL-divergence post-verification demonstrate how verification aids better alignment to the optimal policy and, thus, improves model calibration.
The low rejection rate ensures the computational cost doesn't increase significantly, while the high verification precision underlines the strong utility of the verification step.

\begin{table}[t]
\centering
\small
\setlength\tabcolsep{4pt}
\begin{tabular}{p{3.5cm}|cccc}
\toprule
\textbf{Model}               & \textbf{EM}       & \textbf{F1}       & \textbf{\# T}           & \textbf{\# R}      \\
\midrule
Fixed-base (Retrieval) & 31.6 & 40.4 & 101 & 1 \\
Fixed-sft (Retrieval) & 32.8 & 40.0 & 56 & 1 \\
\midrule
0-shot \modelName & 32.1 & 40.6 & 100 & 0.97 \\
Few-shot \modelName & 28.7 & 37.0 & 93 & 0.79 \\
\midrule
\textbf{Fine-tuned \modelName} & \textbf{37.8} & \textbf{46.0} & \textbf{28} & \textbf{0.48} \\
\bottomrule
\end{tabular}
\caption{Ablation analysis on 2WikiMultihopQA for the need to fine-tune LLMs to incorporate \modelName.}
\label{tab:ablation-finetuning}
\end{table}

\subsubsection{Fine-tuning improves calibration}
\label{sec:ablation-finetuning}

Choosing the right strategy in a 0-shot way is difficult, as models don't surely know what they know and don't know \cite{yin2023largelanguagemodelsknow}.
We analyze the 0-shot \modelName{} strategy planning in Figure~\ref{fig:decision-analysis} and note how the 0-shot model mostly resorts to the most expensive strategy, while fine-tuning helps to learn the patterns between questions and model capabilities.
We also compare the model performance of 0-shot / few-shot \modelName{} with fine-tuned \modelName{} in Table~\ref{tab:ablation-finetuning} for 2WikiMultihopQA.
Clearly, the non-fine-tuned models fail to improve over the fixed strategy baseline, but fine-tuning provides strong performance gains - demonstrating how fine-tuning strongly improves calibration for strategy planning.

\subsubsection{Optimal Policy Upper Bound}

\begin{table}[t]
\centering
\small
\setlength\tabcolsep{4pt}
\begin{tabular}{p{2.3cm}|cc|cc|cc}
\toprule
\textbf{Model}          & \multicolumn{2}{l|}{\textbf{HotpotQA}} & \multicolumn{2}{l|}{\textbf{2WikiQA}} & \multicolumn{2}{l}{\textbf{Musique}} \\
\textbf{}               & \textbf{EM}       & \textbf{F1}       & \textbf{EM}           & \textbf{F1}          & \textbf{EM}       & \textbf{F1}      \\
\midrule
\modelName-base & 36.1 & 47.6 & 37.8 & 46.0 & 10.1 & 19.8 \\
\modelVerifyName & 36.7 & 48.5 & 40.5 & 49.6 & 10.8 & 20.4 \\
\midrule
Upper Bound & 47.0 & 60.5 & 51.3 & 60.2 & 14.2 & 23.0 \\
\quad + $\Delta$ & \textbf{10.3} & \textbf{12.0} & \textbf{10.8} & \textbf{10.6} & \textbf{3.4} & \textbf{2.6} \\
\bottomrule
\end{tabular}
\caption{Empirical upper bounds for possible improvements of \modelName{} using an oracle Decision component with base LLM responses for Execution. $\Delta$ indicates the potential improvement gap.}
\label{tab:theoretical-bound}
\end{table}

We study the upper bound for \modelName{} to motivate possibilities of future improvements.
Specifically, we replace the \modelName's Decision component with the optimal policy and use the corresponding fixed strategy base LLM outputs for Execution.
We compare this upper bound with \modelName{} in Table~\ref{tab:theoretical-bound} with $\Delta$, indicating further potential improvement.
While Musique exhibits a low $\Delta$ of 2-3 F1 points, the larger $\Delta$ of 10-12\% F1 for the other datasets provides promise to further explore strategy planning.

\begin{figure}[t]
    \centering
    \includegraphics[width=\linewidth]{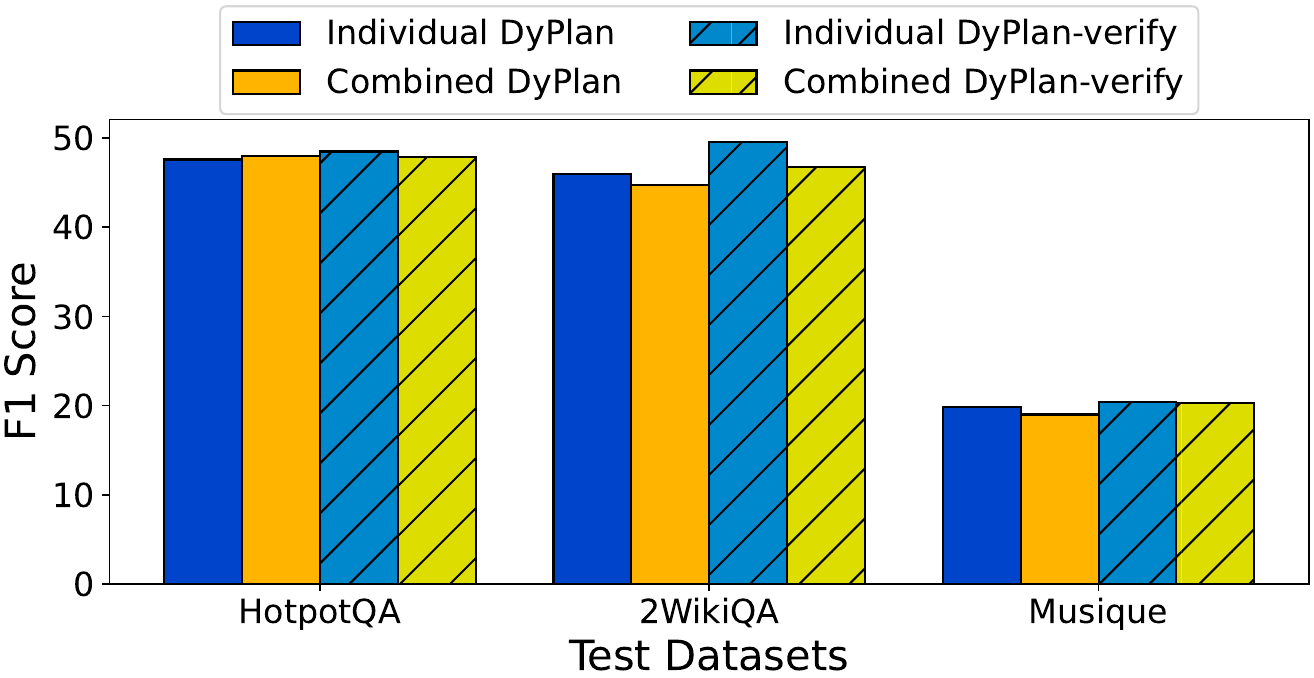}
    \caption{Assessing the generalizability of \modelName{} by comparing the performance (F1 score) of combined-data training with individal-data training.}
    \label{fig:combined-data-ablation}
\end{figure}

\subsection{Generalization Analysis}
\label{sec:combined-data}

To assess the generalization of \modelName, we fine-tune it on combined data from the three benchmark datasets.
To ensure a fair comparison, the combined data comprises 20k datapoints (same as individual data) with equal shares from the three datasets.
We compare the performance of this combined-data fine-tuned model with the individual-data fine-tuned model in Figure~\ref{fig:combined-data-ablation} and note how the performances are nearly similar for both models.
The cost analysis for the combined data model (Table~\ref{tab:combined-data-cost-ablation} in \S~\ref{sec:appendix-combined-data}) reveals at-par levels of computational costs as well.
Thus, this study reveals how the gains provided by \modelName/\modelVerifyName{} are generalizable and not overfitting to a single dataset.

\begin{table}[t]
\centering
\small
\setlength\tabcolsep{4pt}
\begin{tabular}{p{4cm}|cccc}
\toprule
\textbf{Strategy Ordering} & \textbf{EM} & \textbf{F1} & \textbf{\# T} & \textbf{\# R} \\
\midrule
\multicolumn{5}{c}{\modelName} \\
\midrule
Direct, Plan, Reason, Retrieve & 36.1 & 47.6 & 42 & 0.76 \\
Direct, Reason, Plan, Retrieve & 35.4 & 46.8 & 40 & 0.69 \\
Reason, Direct, Plan, Retrieve & 36.1 & 47.7 & 42 & 0.72 \\
\midrule
\multicolumn{5}{c}{\modelVerifyName} \\
\midrule
Direct, Plan, Reason, Retrieve & 36.7 & 48.5 & 53 & 0.78 \\
Direct, Reason, Plan, Retrieve & 36.5 & 48.5 & 51 & 0.72 \\
Reason, Direct, Plan, Retrieve & 37.2 & 48.5 & 58 & 0.76 \\
\bottomrule
\end{tabular}
\caption{Impact of different strategy orderings for \modelName{} on downstream performance and computational costs on the HotpotQA dataset.}
\label{tab:strategy-ordering-sensitivity}
\end{table}

\subsection{Analyzing the Order of Strategies}
\label{sec:strategy-order-analysis}

In this analysis, we study the impact of different strategy ordering $\mathcal{S}$ for \modelName.
Specifically, we consider three different orderings for HotpotQA and show the results in Table~\ref{tab:strategy-ordering-sensitivity}.
Differently ordering the strategies (e.g., Direct, Reason, Plan, Retrieve) can help further reduce the computational cost by 4-7\%, but it can also reduce the performance by 1-2\%.
Similarly, the performance can be optimized by a different ordering (e.g., Reason, Direct, Plan, Retrieve), with improvements upto 1-2\% while incurring additional computational costs upto 10\%.
This puts focus on exploring the choice of the right strategy ordering as a key component for optimization, but we will keep that for future works.

\section{Related Works}

\paragraph{Question Answering:}
Question-answering is a popular task, with wide-spread applications such as document parsing \cite{suvarna-etal-2024-qudselect}, information extraction \cite{parekh-etal-2024-event}, chatbots \cite{chalkidis-etal-2022-lexglue, DBLP:journals/corr/abs-2305-09617}, summarization \cite{fabbri-etal-2022-qafacteval}, as well as great multilingual \cite{parekh-etal-2024-contextual, parekh-etal-2024-speed} and multimodal \cite{DBLP:conf/cvpr/SinghNSJCBPR19, DBLP:journals/corr/abs-2104-06039} applications.
Some prominent benchmarking QA datasets include SQuAD \cite{rajpurkar-etal-2016-squad, rajpurkar-etal-2018-know}, MS MARCO \cite{ms-marco}, TriviaQA \cite{joshi-etal-2017-triviaqa}, and SearchQA \cite{searchqa}.
These datasets are single-hop, i.e., they require simple reasoning to find the answer and are easier to answer.
To develop complex reasoning in models, multi-hop question-answering (MHQA) datasets were developed like HotpotQA \cite{yang-etal-2018-hotpotqa}, 2WikiMultihopQA \cite{ho-etal-2020-constructing}, Compositional Celebrities \cite{press-etal-2023-measuring}, and Musique \cite{trivedi-etal-2022-musique-updated}.
To improve MHQA performance, works explored on improving the reasoning capabilities of LLMs using chain-of-thought \cite{chain-of-thought, cot-2}, auto-CoT \cite{auto-cot}, self-consistency \cite{self-consistency}, tree-of-thought \cite{tree-of-thought}.
Another line of work focused on planning by question decomposition like SelfAsk \cite{press-etal-2023-measuring}, ART \cite{art}, decomposed prompting \cite{decomposed-prompting} or using explicit planners like ReWOO \cite{rewoo}, LLMCompiler \cite{llm-compiler}, stepback \cite{stepback}.
Works also focus on using external knowledge through retrieval like RAG \cite{rag}, Self-RAG \cite{self-rag}, CRAG \cite{crag} with recent works like IRCoT \cite{trivedi-etal-2023-interleaving}, FLARE \cite{jiang-etal-2023-active}, SynCheck \cite{syncheck}, and DRAGIN \cite{su-etal-2024-dragin} exploring dynamic retrieval.
The closest approach to our work is Adaptive RAG \cite{jeong-etal-2024-adaptive} which utilizes a classifier to determine the question complexity and adaptively utilize RAG.
In comparison, our work is more generalized to adapt to any kind of prompt/tool and induces deeper thinking in the LLM itself, while being highly cost-effective at the same time.

\paragraph{Agentic LLMs:}
Recent works have explored LLMs as decision-makers, especially in interactive environments, as agents deciding a policy.
WebGPT \cite{webgpt} utilized LLMs to search the web to answer complex questions.
Some works have also been explored in conversational modeling like BlenderBot \cite{blenderbot}, SimpleTOD \cite{simple-tod}, Tartan \cite{tartan} and robotics like SayCan \cite{saycan} and Inner Monologue \cite{inner-monologue}.
ReAct \cite{react} was one of the earlier systems utilizing natural language thoughts and actions, followed by other works like Reflexion \cite{reflexion} and CAMEL \cite{camel}.

\paragraph{Uncertainty estimation in LLMs:}
With the increasing utilization of LLMs in various reasoning tasks, several works have studied LLM's confidence in its self-knowledge.
\citet{xiao-wang-2021-hallucination} show evidence of model uncertainty with increased hallucinations, while LLM's quantification about its self-knowledge is studied as honesty alignment by \citet{honesty-alignment}.
\citet{llms-know-what-they-know} and \citet{just-ask-for-calibration} discuss how LLMs are generally well-calibrated when for simpler tasks or in the presence of source information.
On the other hand, \citet{llms-taught-what-they-dont-know} and \citet{yin-etal-2023-large} show that LLM calibration about its self-knowledge is not good and explore how fine-tuning can further improve this calibration.

\section{Conclusion and Future Work}
\label{sec:conclusion}

In our work, we introduce the paradigm of dynamic strategy planning for question-answering mimicking human cognitive thinking through \modelName{} with the goal of reducing inference computational costs and improving model performance.
By adding verification and self-correction using \modelVerifyName, we further enhance the model output quality.
Through experimentation on three MHQA datasets, we show strong efficacy and improved performance using our techniques.
Our analyses and empirical bounds provide promise for further improvements.
Incorporating partial thinking and integrating dynamic tool usage can be explored to further improve \modelName.
Utilizing alignment-based fine-tuning can further improve the model's effectiveness.

\section*{Limitations}

We present a prototype for our technique \modelName{} to selectively choose strategies in our work. We haven't evaluated it extensively on all possible strategies, tools, and models and we leave it for future work.
Our technique is not restricted to question-answering and is generalizable to other tasks as well. But in this work, we only show experiments and results on question-answering.
We haven't optimized our technique or explored changing the hyper-parameters or the prompt. It might be possible to improve the model by further engineering, but again, we leave it up to future work.
For fine-tuning, we limit ourselves to LoRA and smaller models (8B) only owing to budget constraints. Full fine-tuning and exploring larger LLMs might be faster and better and can be explored in future works.

\section*{Ethical Considerations}

We utilize LLMs to partially correct and rewrite parts of our paper.
Since we work with generative models, there's little control on the text/tokens. It is possible that the model can generate spurious or unsafe content and we haven't evaluated our trained models for it.
In general, fine-tuning has been prone to reducing the robustness of LLMs for other tasks/skills or introducing additional biases due to spurious patterns in training. We haven't evaluated the models for robustness or general safety.
Finally, our work promotes using less retrieval in favor of reducing inference generation costs. Previous works have found an inverse correlation between hallucinations and knowledge grounding in external documents. So, our work can induce more hallucinations at the cost of reducing inference costs, and this should be taken into consideration before using our work.

\bibliography{anthology, custom}

\clearpage

\appendix

\section{Additional Implementation Details}
\label{sec:hyperparams}

In this section, we provide additional details about our implementation and hyperparameters of each technique.

\subsection{General Implementation Details}

All of our experiments were conducted on an NVIDIA RTX A100 machine with support for 8 GPUs.
Fine-tuning runs took about 6-18 hours to complete using distributed training on 4 GPUs.
Inference was faster and would be completed in 1-2 hours on a single GPU.
Our base LLM for all experiments was Llama3-8B-instruct \cite{llama3}, specifically its Huggingface release.\footnote{\url{https://huggingface.co/meta-llama/Meta-Llama-3-8B-Instruct}}
We average the main results for most techniques over three runs.
Final inference was run with temperature = 0.4 leading to low variance in model performance.

\subsection{Fixed Strategy Implementations}

We self-implemented the simple \textit{Direct} strategy.
We utilized the codebase\footnote{\url{https://github.com/oneal2000/DRAGIN}} of DRAGIN \cite{su-etal-2024-dragin} to implement the \textit{Chain-of-Thought} (CoT) \cite{chain-of-thought} and \textit{RAG} \cite{rag} strategies.
We utilized a BM25 retrieval system indexed on the entire Wikipedia and capable of retrieving intermediate excerpts of length 200 based on the query.
We provide the top three passages as the retrieved passages for RAG.
For SelfAsk \cite{press-etal-2023-measuring}, we utilized their original codebase.\footnote{\url{https://github.com/ofirpress/self-ask}}
If any of the strategy inferences weren't able to provide their answer within the max generation length limit, we used force-decoding with a preset prefix ``Final answer:" to get the final answer.
Other specific hyperparameters are provided for each strategy in Tables~\ref{tab:direct-fixed-hyper},~\ref{tab:cot-fixed-hyper},~\ref{tab:plan-fixed-hyper}, and ~\ref{tab:rag-fixed-hyper}.

\begin{table}[h]
    \centering
    \small
    \begin{tabular}{lr}
        \toprule
        \textbf{\# In-context Examples} & 8 \\
        \textbf{Max Generation Length} & 100 \\
        \bottomrule
    \end{tabular}
    \caption{Hyper-parameters for Fixed Strategy Implementation for Direct strategy. Here, \# = number of.}
    \label{tab:direct-fixed-hyper}
\end{table}

\begin{table}[h]
    \centering
    \small
    \begin{tabular}{lr}
        \toprule
        \textbf{\# In-context Examples} & 8 \\
        \textbf{Max Generation Length} & 200 \\
        \bottomrule
    \end{tabular}
    \caption{Hyper-parameters for Fixed Strategy Implementation for Direct strategy. Here, \# = number of.}
    \label{tab:cot-fixed-hyper}
\end{table}

\begin{table}[h]
    \centering
    \small
    \begin{tabular}{lr}
        \toprule
        \textbf{\# In-context Examples} & 4 \\
        \textbf{Max Generation Length} & 200 \\
        \bottomrule
    \end{tabular}
    \caption{Hyper-parameters for Fixed Strategy Implementation for Direct strategy. Here, \# = number of.}
    \label{tab:plan-fixed-hyper}
\end{table}

\begin{table}[h]
    \centering
    \small
    \begin{tabular}{lr}
        \toprule
        \textbf{\# In-context Examples} & 8 \\
        \textbf{Retriever} & BM25 \\
        \textbf{\# Retrievals} & 3 \\
        \textbf{Max Generation Length} & 200 \\
        \bottomrule
    \end{tabular}
    \caption{Hyper-parameters for Fixed Strategy Implementation for Direct strategy. Here, \# = number of.}
    \label{tab:rag-fixed-hyper}
\end{table}

\subsection{\modelName{} Implementation}

We describe the prompts and multi-turn setting of our model \modelName{} in \S~\ref{sec:methodology}.
We provide specific hyperparameters for the non-fine-tuned version of \modelName{} and \modelVerifyName{} in Table~\ref{tab:dyplan-hyper}.
We utilize the hierarchy order of Direct < Plan < Reason < Retrieval for the Decision component.
For RAG strategy selection, we provide the retrieved passages as part of the Execution prompt.
We set the number of Decision-Execution-Verification rounds for \modelVerifyName{} to 2.

\begin{table}[h]
    \centering
    \small
    \begin{tabular}{lr}
        \toprule
        \textbf{\# In-context Examples} & 0-4 \\
        \textbf{Retriever} & BM25 \\
        \textbf{\# Retrievals} & 3 \\
        \textbf{Max Generation Length for Decision} & 10 \\
        \textbf{Max Generation Length for Execution} & 200 \\
        \textbf{Max Generation Length for Verification} & 10 \\
        \textbf{Numbers of Rounds} & 2 \\
        \bottomrule
    \end{tabular}
    \caption{Hyper-parameters for \modelName{} and \modelVerifyName. Here, \# = number of.}
    \label{tab:dyplan-hyper}
\end{table}

\subsection{Fine-Tuning Details}

We utilize LoRA \cite{lora} for fine-tuning the base LLM for fixed strategy and \modelName.
We utilize the LLaMa-Factory \cite{zheng-etal-2024-llamafactory} and their codebase\footnote{\url{https://github.com/hiyouga/LLaMA-Factory}} for the fine-tuning and inference.
We provide the hyperparameters for this tuning in Table~\ref{tab:fine-tuning-hyper}.

\begin{table}[h]
    \centering
    \small
    \begin{tabular}{lr}
        \toprule
        \textbf{\# In-context Examples} & 0 \\
        \textbf{LoRA rank} & 32 \\
        \textbf{LoRA target} & All \\
        \textbf{Train Datasize} & 20,000 \\
        \textbf{Learning Rate} & 1e-5 \\
        \textbf{Warmup Ratio} & 0.1 \\
        \textbf{\# Epochs} & 4 \\
        \textbf{Save Steps} & 250 \\
        \textbf{Train Batch size} & 16 \\
        \textbf{Inference Batch size} & 32 \\
        \bottomrule
    \end{tabular}
    \caption{Hyper-parameters for fine-tuning the base LLM. Here, \# = number of.}
    \label{tab:fine-tuning-hyper}
\end{table}

\subsection{Classifier Implementation}

As a baseline, we train a multi-class classifier with each strategy as a separate class to select an appropriate strategy based on the question.
We experimented with utilizing binary classifiers for each strategy but the multi-class classifier performed better.
We utilize the codebase\footnote{\url{https://github.com/google-research/xtreme}} from XTREME \cite{xtreme} to implement the classifiers.
We utilize RoBerta-large \cite{roberta} as the base model.
We provide additional details about the hyperparameters in Table~\ref{tab:classifier-hyper}.

\begin{table}[h]
    \centering
    \small
    \begin{tabular}{lr}
        \toprule
        \textbf{Base Model} & RoBerta-large \\
        \textbf{Max length} & 256 \\
        \textbf{Train Batch size} & 32 \\
        \textbf{Train datasize} & 20,000 \\
        \textbf{Learning Rate} & 1e-5 \\
        \textbf{Weight Decay} & 0 \\
        \textbf{Warmup Steps} & 0 \\
        \textbf{\# Epochs} & 10 \\
        \textbf{Save Steps} & 50 \\
        \textbf{Adam Epsilon} & 1e-8 \\
        \textbf{Max Gradient Norm} & 1.0 \\
        \bottomrule
    \end{tabular}
    \caption{Hyper-parameters for fine-tuning the base LLM. Here, \# = number of.}
    \label{tab:classifier-hyper}
\end{table}

\subsection{Majority Ensemble Implementation}

We implement a simple majority ensemble wherein we utilize the final answers from the fixed strategy models and aggregate them using a majority function.
In case of a tie, we choose the final answer of the better strategy in the hierarchy.
In 2-3 strategy cases, this leads to aligning with the best-fixed strategy method itself.

\subsection{ReAct Implementation}

As a reference for costs, we also included a baseline for ReAct \cite{react}.
We utilize their original codebase\footnote{\url{https://github.com/ysymyth/ReAct}} for the implementation.
Utilizing the Instruct version of Llama3-8B didn't work as well, instead we utilize the non-instruct-tuned version of this model Llama3-8B\footnote{\url{https://huggingface.co/meta-llama/Meta-Llama-3-8B}} for this baseline.
We utilize six in-context examples for the prompt.
Additionally, to keep a fair comparison and reduce token generation costs, we do forced decoding stopping for the keywords of "Thought:", "Action:" or "Observation".
This avoids any unnecessary token generations or when the model starts to repeat itself.
We notice that this model works well with larger LLMs, but the planning and performance are poor with smaller LLMs.

\subsection{DRAGIN Implementation}

As a reference for costs, we also included a baseline for ReAct \cite{su-etal-2024-dragin} implemented using their original implementation codebase.~\footnote{\url{https://github.com/oneal2000/DRAGIN}}
We provide specific hyperparameters of this model in Table~\ref{tab:dragin-hyper}.

\begin{table}[h]
    \centering
    \small
    \begin{tabular}{lr}
        \toprule
        \textbf{\# In-context Examples} & 8 \\
        \textbf{Retriever} & BM25 \\
        \textbf{\# Retrievals} & 3 \\
        \textbf{\# Retrieval Keep Top-k} & 25 \\
        \textbf{Max Generation Length} & 200 \\
        \textbf{Hallucination Threshold} & 1.0 \\
        \textbf{Query Formulation} & real-words \\
        \textbf{Check Real Words} & true \\
        \bottomrule
    \end{tabular}
    \caption{Hyper-parameters for Fixed Strategy Implementation for Direct strategy. Here, \# = number of.}
    \label{tab:dragin-hyper}
\end{table}

\section{Additional Experimental Results}
\label{sec:appendix-expts}

\subsection{Hierarchy Violations for other datasets}
\label{sec:appendix-hierarchy-violations}

In \S~\ref{sec:methodology-motivation}, we motivated how strategy selection acts as an ensemble for the Direct-Reason strategy combination.
Here, we provide more evidence to support this claim across four strategies - Direct, Plan, Reason, and Retrieval - and multiple datasets.

\paragraph{General Hierarchy:}
For the four strategies mentioned above, a general hierarchy we assume is Direct < Plan < Reason < Retrieval.
If the model knows the answer directly, then it should be able to plan/reason to provide the answer. Thus, Direct is the lowest in this hierarchy.
Comparing Plan and Reason - we assume Plan is a special kind of reasoning with a specific focus on breaking the question into atomic units.
On the other hand, there are several questions like ``\textit{Who was the actor who starred in an Avengers movie and has three children?}" where breaking into atomic questions will not help to answer the question.
Thus, we assume Plan < Reason.
Finally, Retrieval brings in additional external information compared to Reason ranking Reason < Retrieval.

\paragraph{Hierarchy Violations:}
In an ideal world, the LLM should follow this hierarchy, and we should simply use Retrieval all the time to optimize model performance.
However, owing to various reasons like non-relevant retrievals, incorrect reasoning, rote learning, and spurious generations, this hierarchy is not maintained.
We call these special cases as \textit{hierarchy violations}.

\begin{figure}[t]
    \centering
    \includegraphics[width=\linewidth]{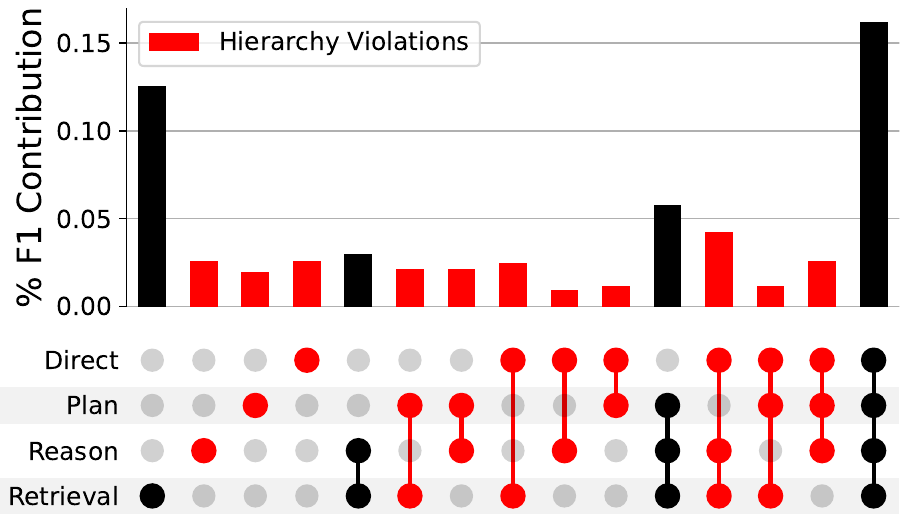}
    \caption{Breaking the contribution of each strategy combination to HotpotQA model performance. A strategy's inclusion in the set is indicated by the colored dot. \textcolor{red}{Red} dots and bars indicate the hierarchy violations.}
    \label{fig:hotpotqa-venn}
\end{figure}

\begin{figure}[t]
    \centering
    \includegraphics[width=\linewidth]{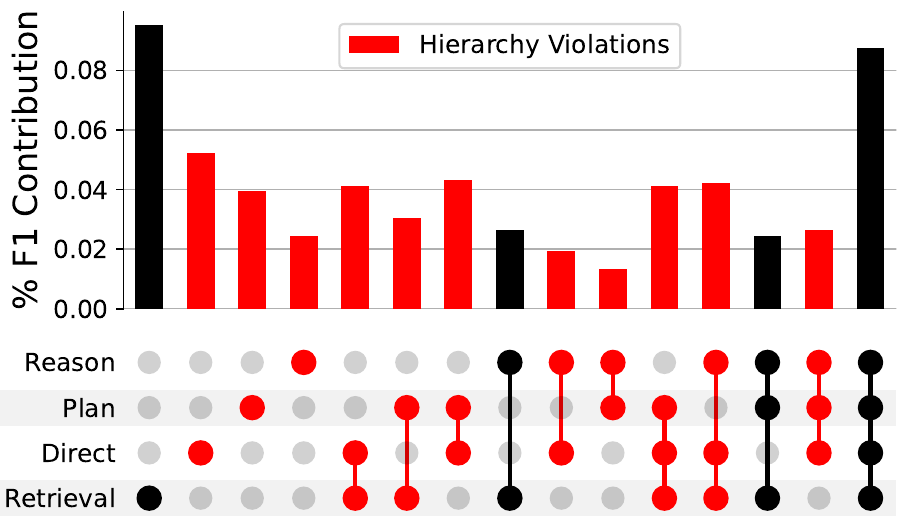}
    \caption{Breaking the contribution of each strategy combination to 2WikiMultihopQA model performance. A strategy's inclusion in the set is indicated by the colored dot. \textcolor{red}{Red} dots and bars indicate the hierarchy violations.}
    \label{fig:2wiki-venn}
\end{figure}

\begin{figure}[t]
    \centering
    \includegraphics[width=\linewidth]{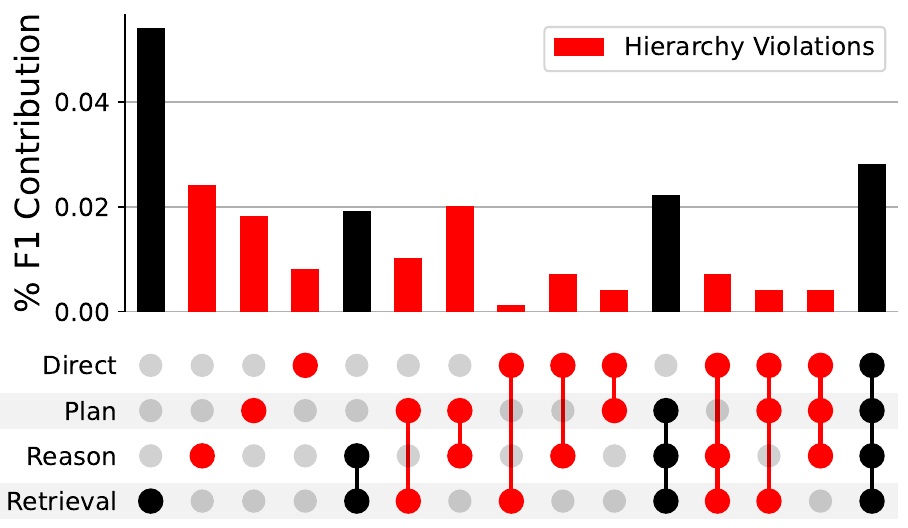}
    \caption{Breaking the contribution of each strategy combination to Musique model performance. A strategy's inclusion in the set is indicated by the colored dot. \textcolor{red}{Red} dots and bars indicate the hierarchy violations.}
    \label{fig:musique-venn}
\end{figure}

\paragraph{Quantifying Violations:}
Similar to the study in \S~\ref{sec:methodology-motivation}, we quantify the F1 performance contribution of the hierarchy violations for all the four strategies using Llama3-8B-Instruct for the three datasets of HotpotQA, 2WikiMultihopQA and Musique in Figures~\ref{fig:hotpotqa-venn}, \ref{fig:2wiki-venn} and \ref{fig:musique-venn}.
These upset plots are a way of visualizing Venn diagrams, wherein each column is a unique combination of strategies, and the bar heights indicate its F1 contribution.
The colored dots (black/red) indicate the presence of the corresponding strategy in the strategy combination set, while the grey dots indicate the absence.
For example, in Figure~\ref{fig:hotpotqa-venn}, the first bar indicates that there are about 9.5\% questions that only Retrieval can correctly answer while all other methods fail.
Similarly, the second bar indicates that more than 5\% questions can only be answered by Direct and no other strategy.
To distinctively show the hierarchy violations, we color-code them in \textcolor{red}{red} in these plots. 

\paragraph{Results:}
Similar to our original findings, we find a significant portion of performance contributions can potentially be attributed to hierarchy violation patterns.
Specifically, they account for approximate F1 scores of 23.5\%, 37\%, and 9\% for HotpotQA, 2WikiMultihopQA, and Musique, respectively.
We provide some qualitative examples to back this finding more in \S~\ref{sec:appendix-strategy-hierarchy-examples}
Such high contributions from hierarchy violations indicate that the underlying hierarchy is weak, and ensembling can yield better-combined performance.
To this end, our technique \modelName{} can provide promise by utilizing dynamic strategy planning not only to reduce costs but also to improve model performance.

\subsection{Fine-tuning ablation Analysis}
\label{sec:appendix-ablation-finetuning}

We provided basic ablation analysis highlighting how fine-tuning aids LLMs to be better calibrated to utilize \modelName{} in \S~\ref{sec:ablation-finetuning}.
Here we provide additional details about the experimental setup along with additional results on other datasets.

\paragraph{Non fine-tuned \modelName:}
For zero-shot model, we prompt the base LLM for the Decision component (i.e. to select the preferred strategy) in a zero-shot fashion.
Based on the chosen strategy, the base LLM output from the fixed strategy run is used as the Execution component output.
For the few-shot model, we utilize four few-shot examples for each strategy (16 in total).
If the model outputs a strategy not defined in the list of provided strategies, we map it to the retrieval strategy (as that's the best-preferred strategy in terms of model performance).

\paragraph{Results:}
We provided results for this study for the 2WikiMultihopQA dataset in Table~\ref{tab:ablation-finetuning}.
Here, we also provide similar comparisons on HotpotQA and Musique datasets in Tables~\ref{tab:hotpotqa-ablation-finetuning} and \ref{tab:musique-ablation-finetuning}, respectively.
Across all the datasets, we can notice the sub-optimal performance of non-fine-tuned LLM runs with \modelName.
The zero-shot model mostly selects retrieval, while the few-shot model selects other strategies, but it's not well-calibrated.
The calibration is poorer for few-shot \modelName as the model gets heavily influenced by the in-context examples.
In conclusion, we demonstrate how base LLMs by default are not calibrated well to utilize \modelName{}, underlining the need for fine-tuning LLMs.

\begin{table}[t]
\centering
\small
\setlength\tabcolsep{4pt}
\begin{tabular}{p{3.5cm}|cccc}
\toprule
\textbf{Model}               & \textbf{EM}       & \textbf{F1}       & \textbf{\# T}           & \textbf{\# R}      \\
\midrule
Fixed-base (Retrieval) & 36.1 & 47.9 & 185 & 1 \\
Fixed-sft (Retrieval) & \textbf{36.8} & \textbf{48.6} & 53 & 1 \\
\midrule
0-shot \modelName & 35.3 & 46.6 & 179 & 0.92 \\
Few-shot \modelName & 33.1 & 44.1 & 167 & 0.79 \\
\midrule
\textbf{Fine-tuned \modelName} & 36.1 & 47.6 & \textbf{42} & \textbf{0.76} \\
\bottomrule
\end{tabular}
\caption{Ablation analysis on HotpotQA for the need to fine-tune LLMs to incorporate \modelName.}
\label{tab:hotpotqa-ablation-finetuning}
\end{table}

\begin{table}[t]
\centering
\small
\setlength\tabcolsep{4pt}
\begin{tabular}{p{3.5cm}|cccc}
\toprule
\textbf{Model}               & \textbf{EM}       & \textbf{F1}       & \textbf{\# T}           & \textbf{\# R}      \\
\midrule
Fixed-base (Retrieval) & 9.6 & 18.0 & 187 & 1 \\
Fixed-sft (Retrieval) & 9.3 & 18.4 & 88 & 1 \\
\midrule
0-shot \modelName & 8.7 & 17.2 & 183 & 0.95 \\
Few-shot \modelName & 7.3 & 15.9 & 175 & \textbf{0.85} \\
\midrule
\textbf{Fine-tuned \modelName} & \textbf{10.1} & \textbf{19.8} & \textbf{65} & 0.98 \\
\bottomrule
\end{tabular}
\caption{Ablation analysis on Musique for the need to fine-tune LLMs to incorporate \modelName.}
\label{tab:musique-ablation-finetuning}
\end{table}

\begin{figure}[t]
    \centering
    \includegraphics[width=\linewidth]{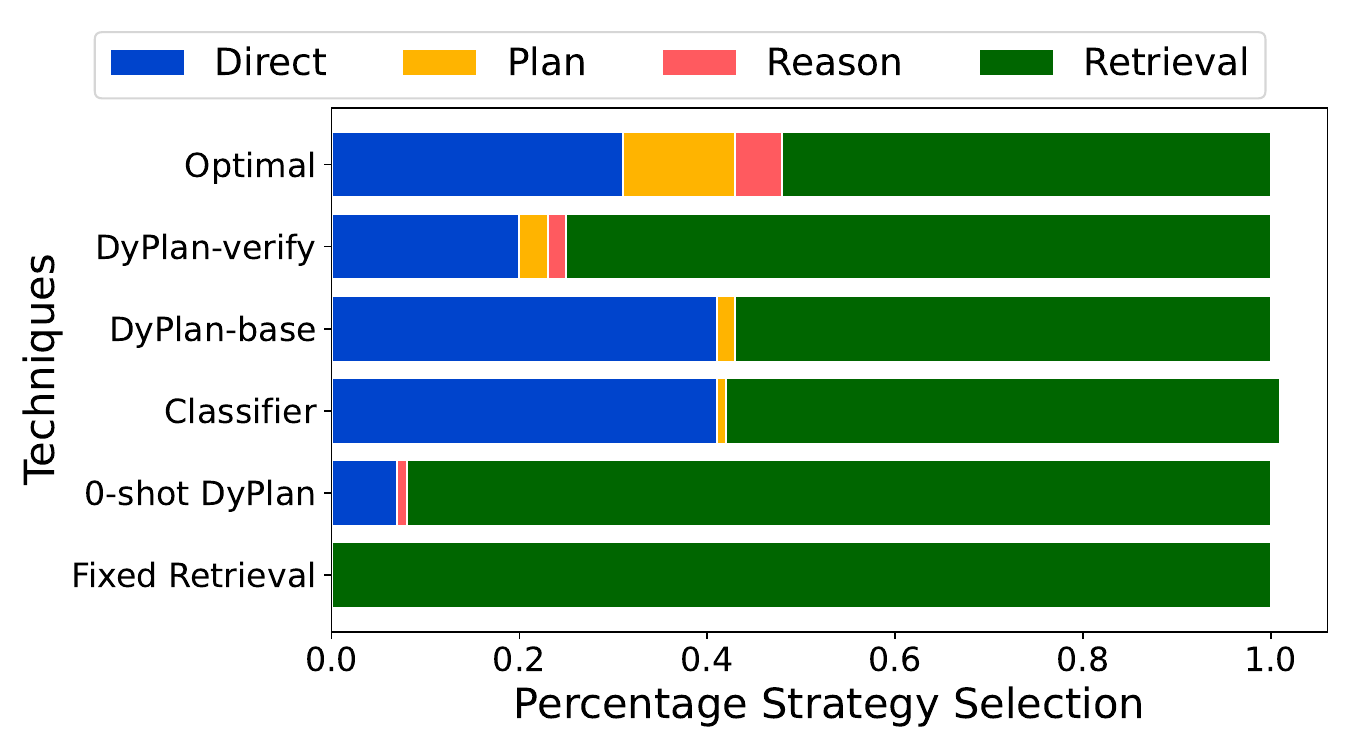}
    \caption{Comparing the strategy usage distribution of various techniques with the optimal policy distribution for HotpotQA.}
    \label{fig:decision-analysis-hotpotqa}
\end{figure}

\begin{figure}[t]
    \centering
    \includegraphics[width=\linewidth]{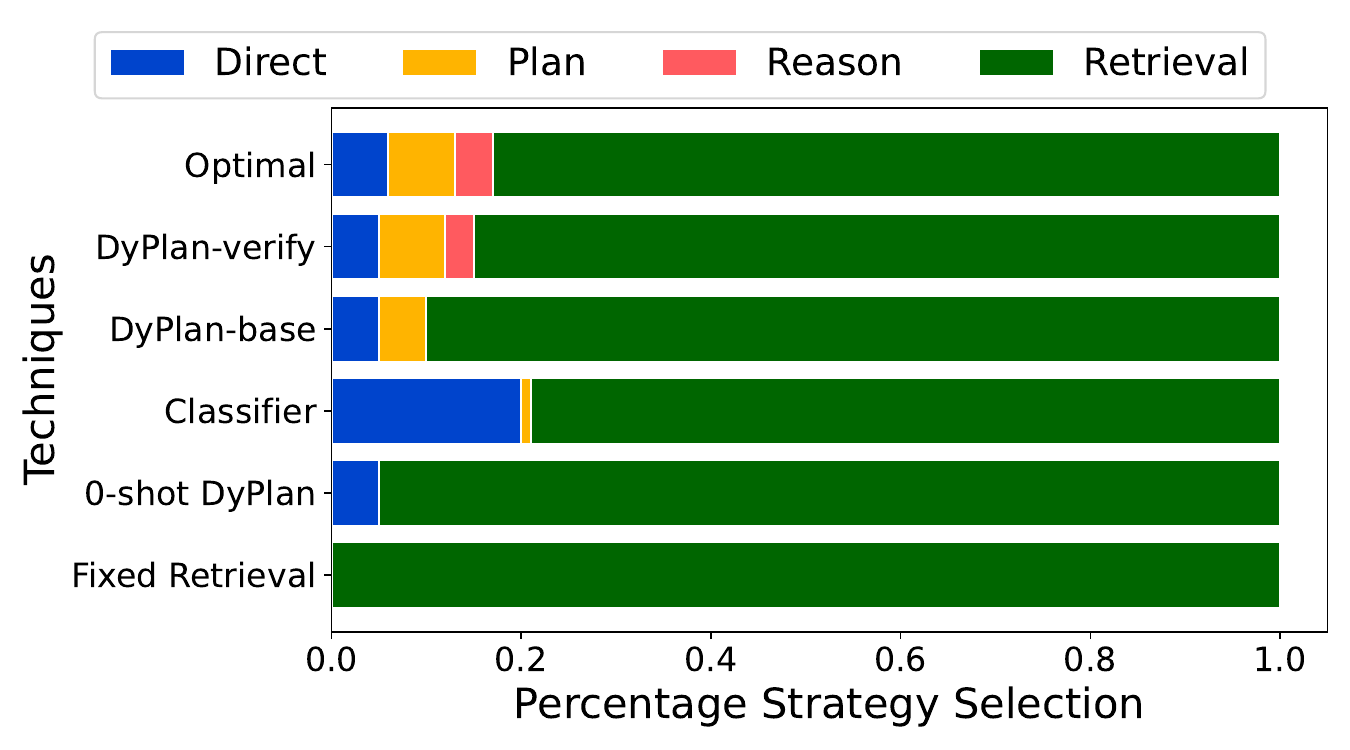}
    \caption{Comparing the strategy usage distribution of various techniques with the optimal policy distribution for Musique.}
    \label{fig:decision-analysis-musique}
\end{figure}

\subsection{Decision-making of \modelName}
\label{sec:appendix-decision-making}

We discussed how \modelName{} helps to better calibrate decision-making in terms of strategy selection for the 2WikiMultihopQA dataset in \S~\ref{sec:decision-making-analysis}.
Here, we show similar analysis for the other datasets of HotpotQA and Musique in Figures~\ref{fig:decision-analysis-hotpotqa} and \ref{fig:decision-analysis-musique}.
Similar to our earlier findings, we notice the \modelName{} helps the strategy usage to be more similar to the optimal policy, in turn helping to improve model performance.
We notice for HotpotQA, \modelName{} is similar to the external classifier.
\modelVerifyName, on the other hand, is strongly closer to the optimal policy for both HotpotQA and Musique.

\subsection{Combined Data Fine-tuning}
\label{sec:appendix-combined-data}

In \S~\ref{sec:combined-data}, we compared and discussed the performance difference for models fine-tuned on individual datasets relative to models fine-tuned on a single combined dataset.
We provide the complete table with the performance numbers in Table~\ref{tab:combined-data-ablation}.
We also compare the costs of these models in Table~\ref{tab:combined-data-cost-ablation}.
We observe that the costs for \modelName{} with combined data are slightly more than the individual models.
On the contrary, the costs for a combined model for \modelVerifyName{} are lesser than the individual model fine-tuning.
Overall, the range of differences is quite small, and the combined model costs less than the best baseline as well.

\begin{table}[t]
\centering
\small
\setlength\tabcolsep{4pt}
\begin{tabular}{p{2.3cm}|cc|cc|cc}
\toprule
\textbf{Model}          & \multicolumn{2}{l|}{\textbf{HotpotQA}} & \multicolumn{2}{l|}{\textbf{2WikiQA}} & \multicolumn{2}{l}{\textbf{Musique}} \\
\textbf{}               & \textbf{EM}       & \textbf{F1}       & \textbf{EM}           & \textbf{F1}          & \textbf{EM}       & \textbf{F1}      \\
\midrule
\multicolumn{7}{c}{\modelName-base} \\
\midrule
Individual & 36.1 & 47.6 & 37.8 & 46.0 & 10.1 & 19.8 \\
Combined & 36.5 & 48.0 & 36.2 & 44.7 & 9.8 & 19.0 \\
\quad - $\Delta$ & -0.4 & -0.4 & 1.6 & 1.3 & 0.3 & 0.8 \\
\midrule
\multicolumn{7}{c}{\modelVerifyName} \\
\midrule
Individual & 36.7 & 48.5 & 40.5 & 49.6 & 10.8 & 20.4 \\
Combined & 36.1 & 47.9 & 38.1 & 46.8 & 10.6 & 20.3 \\
\quad - $\Delta$ & 0.6 & 0.6 & 2.4 & 2.8 & 0.2 & 0.1 \\
\bottomrule
\end{tabular}
\caption{Generalization analysis comparing model performance of LLM fine-tuned on a single combined dataset v/s individual datasets.}
\label{tab:combined-data-ablation}
\end{table}

\begin{table}[t]
\centering
\small
\setlength\tabcolsep{4pt}
\begin{tabular}{p{2.3cm}|cc|cc|cc}
\toprule
\textbf{Model}          & \multicolumn{2}{l|}{\textbf{HotpotQA}} & \multicolumn{2}{l|}{\textbf{2WikiQA}} & \multicolumn{2}{l}{\textbf{Musique}} \\
\textbf{}               & \textbf{\# T}       & \textbf{\# R}       & \textbf{\# T}           & \textbf{\# R}          & \textbf{\# T}       & \textbf{\# R}      \\
\midrule
\multicolumn{7}{c}{\modelName-base} \\
\midrule
Individual & 42 & 0.76 & 28 & 0.48 & 65 & 0.98 \\
Combined & 44 & 0.76 & 31 & 0.58 & 71 & 0.98 \\
\quad - $\Delta$ & 2 & 0 & 3 & 0.1 & 6 & 0 \\
\midrule
\multicolumn{7}{c}{\modelVerifyName} \\
\midrule
Individual & 53 & 0.79 & 45 & 0.65 & 77 & 0.99 \\
Combined & 53 & 0.77 & 44 & 0.54 & 65 & 0.97 \\
\quad - $\Delta$ & 0 & -0.02 & -1 & -0.11 & -12 & -0.02 \\
\bottomrule
\end{tabular}
\caption{Generalizability analysis comparing model performance of LLM fine-tuned on a single combined dataset v/s individual datasets. Here, 2WikiQA = 2WikiMultihopQA}
\label{tab:combined-data-cost-ablation}
\end{table}

\subsection{Generalization with other LLMs}

\begin{table}[t]
\centering
\small
\setlength\tabcolsep{4pt}
\begin{tabular}{p{2.3cm}|cccc}
\toprule
\textbf{Model} & \textbf{EM} & \textbf{F1} & \textbf{\# T} & \textbf{\# R} \\
\midrule
Fixed-sft Direct & 41.6 & 44.2 & 11 & 0 \\
Fixed-sft Reason & 43.6 & 47.7 & 91 & 0 \\
Fixed-sft Plan & 41.4 & 44.7 & 106 & 0 \\
Fixed-sft Retrieval & 47.8 & 52.0 & 99 & 1 \\
\midrule
\modelName-base & 49.1 & 53.6 & 37 & 0.55 \\
\modelVerifyName{} & 49.9 & 54.8 & 55 & 0.59 \\
\bottomrule
\end{tabular}
\caption{Benchmarking model performance using \modelName{} for Mistral-7B model on 2WikiMultihopQA dataset.}
\label{tab:mistral-results}
\end{table}

To validate the compatibility of \modelName{} with other LLMs, we conduct a small study utilizing \modelName{} and \modelVerifyName{} for Mistral-7B \cite{mistral_7b} model.\footnote{\url{https://huggingface.co/mistralai/Mistral-7B-v0.3}}
We provide the experimental results for this model in Table~\ref{tab:mistral-results} for the 2WikiMultihopQA dataset.
The results demonstrate how \modelName{} brings about a 3-5\% boost in model performance and a 41-63\% reduction in token and retrieval costs.
Overall, this provides evidence for the generalizability of \modelName{} across different LLMs.

\section{Qualitative Studies}

\begin{table*}
    \centering
    \small
    \begin{tabular}{p{2.7cm}|p{4.8cm}|p{5.2cm}|p{1.5cm}}
        \toprule
        \textbf{Question} & \textbf{\modelName{} Output} & \textbf{Fixed-sft Retrieval Output} & \textbf{Comments} \\
        \midrule
        Where was the father of Alexandre Gaydamak born? & \textit{Decision}: Direct \linebreak \textit{Execution}: Final answer: "Russia"  & [1] ... [2] ... [3] ... Alexandre Gaydamak's father, Arcadi Gaydamak, was born in France. Final answer: "France" & {\textcolor{darkgreen}{Efficient \& Correct}} \\ \hline
        Who is younger, Lo Hsiao-Ting or Basil Hoffman? & \textit{Decision}: Direct \linebreak \textit{Execution}: Final answer: "Lo Hsiao-Ting" & [1] ... [2] ... [3] ...  Lo Hsiao-Ting was born on 1985-09-29. Basil Hoffman was born on 1938-01-18. Thus, Basil Hoffman is older than Lo Hsiao-Ting. Final answer: "Basil Hoffman" & {\textcolor{darkgreen}{Efficient and Correct}} \\ \hline
        Who lived longer, Csaba Pálinkás or Henry Scheffé? & \textit{Decision}: Plan \linebreak \textit{Execution}: Follow up:  How old was Csaba Pálinkás when he died? Intermediate answer: Csaba Pálinkás was 64 years old when he died. Follow up: How old was Henry Scheffé when he died? Intermediate answer: Henry Scheffé was 72 years old when he died. Final answer: "Henry Scheffé" & [1] ... [2] ... [3] ... Henry Scheffé was born on April 11, 1907, and died on July 5, 1977. Csaba Pálinkás was born on 1936 and died on 2017. Therefore, Henry Scheffé lived for 70 years, and Csaba Pálinkás lived for 81 years. Csaba Pálinkás lived longer. Final answer: "Csaba Pálinkás" & {\textcolor{darkgreen}{Correct by self-knowledge}} \\ \hline
        Where was the father of Eugene Fubini born? & \textit{Decision}: CoT \linebreak \textit{Execution}: Eugene Fubini's father, Guido Fubini, was born in Italy. Final answer: "Italy" & [1] ... [2] ... [3] ... Eugene Fubini's father was Guido Fubini. Guido Fubini was a mathematician. There is no information provided about Guido Fubini's birthplace. Final answer: "unknown" & {\textcolor{darkgreen}{Efficient and correct by self-knowledge}} \\ \hline
        Who is the maternal grandfather of Amadeus Vii, Count Of Savoy? & \textit{Decision}: RAG \linebreak \textit{Execution}: Amadeus VII, Count of Savoy's mother is Bonne of Bourbon. Bonne of Bourbon's father is Peter I, Count of Bourbon. Thus, Amadeus VII, Count of Savoy's maternal grandfather is Peter I, Count of Bourbon. Final answer: "Peter I, Count of Bourbon" & [1] ... [2] ... [3] ...  Amadeus VII, Count of Savoy was the son of Amadeus VI, Count of Savoy and Bonne of Bourbon. Bonne of Bourbon was the daughter of John, Duke of Berry and Joan of Auvergne. Therefore, the maternal grandfather of Amadeus VII, Count of Savoy is John, Duke of Berry. Final answer: "John, Duke of Berry" & {\textcolor{darkgreen}{Correct by better Reasoning}} \\
        \midrule
        Which film has the director who died earlier, Captain Kidd And The Slave Girl or Relax Freddie? & \textit{Decision}: Direct \linebreak \textit{Execution}: Final answer: "Captain Kidd And The Slave Girl" & [1] ... [2] ... [3] ...  Captain Kidd And The Slave Girl was released in 1952. Relax Freddie was released in 1971. Therefore, Relax Freddie was released later. The director of Relax Freddie is unknown. The director of Captain Kidd And The Slave Girl is Lew Landers. Lew Landers died in 1993. Therefore, Lew Landers died earlier. Final answer: "Captain Kidd And The Slave Girl" & {\textcolor{darkgreen}{More efficient}} \\ \hline
        Where was the performer of song I See You (Theme From Avatar) born? & \textit{Decision}: Reason \linebreak \textit{Execution}: The song I See You (Theme From Avatar) was performed by Leona Lewis. Leona Lewis was born in Highbury, London, England. Final answer: "Highbury, London, England" & [1] ... [2] ... [3] ... Leona Lewis, the performer of the song I See You (Theme From Avatar), was born in London, England. Final answer: \"London, England\" & {\textcolor{darkgreen}{Efficient by self-knowledge}} \\
        \bottomrule
    \end{tabular}
    \caption{Qualitative analysis depicting cases where \modelName{} improves over Fixed-sft Retrieval baseline.}
    \label{tab:dyplan-vs-fixed-sft-retrieval}
\end{table*}

\subsection{Qualitative Examples for \modelName}
\label{sec:dyplan-qualitative-examples}

We provide some qualitative examples highlighting the cases where \modelName{} provides stronger model performance and better efficiency compared to the best baseline of Fixed-sft Retrieval in Table~\ref{tab:dyplan-vs-fixed-sft-retrieval}.
We also provide corresponding comments to indicate how \modelName{} is better.
Initial examples demonstrate cases wherein \modelName{} is more efficient as well as correct.
Some of these examples are also when both the methods use RAG - which throws light on improved reasoning ability by \modelName{} training.
At the bottom, we show examples where both give the right answer, but \modelName{} is more efficient.

\begin{table*}
    \centering
    \small
    \begin{tabular}{p{2.7cm}|p{5cm}|p{7cm}}
        \toprule
        \textbf{Question} & \textbf{First Round Output} & \textbf{Second Round Output} \\
        \midrule
        Which film has the director who died earlier, Toen 'T Licht Verdween or Boum Sur Paris? & \textit{Decision}: Direct \linebreak \textit{Execution}: Final answer: "Toen 'T Licht Verdween" \linebreak \textit{Verification}: No & \textit{Decision}: Plan \linebreak \textit{Execution}: Follow up:  Who is the director of Toen 'T Licht Verdween? Intermediate Answer: The director of Toen 'T Licht Verdween is Fons Rademakers. Follow up: When did Fons Rademakers die? Intermediate Answer: Fons Rademakers died on February 22, 2001. Follow up: Who is the director of Boum Sur Paris? Intermediate Answer: The director of Boum Sur Paris is Henri Lepage. Follow up: When did Henri Lepage die? Intermediate Answer: Henri Lepage died on October 14, 1991. Final answer: "Boum Sur Paris" \\ \hline
        Are both directors of films The Princess Bride (Film) and I Never Sang For My Father from the same country? & \textit{Decision}: Direct \linebreak \textit{Execution}: Final answer: "no" \linebreak \textit{Verification}: No & \textit{Decision}: CoT \linebreak \textit{Execution}: The Princess Bride (film) was directed by Rob Reiner. I Never Sang For My Father was directed by John Erman. Rob Reiner's nationality is United States. John Erman's nationality is United States. Thus, they are from the same country. Final answer: "yes" \\ \hline
        Where was the director of film Lettre Ouverte born? & \textit{Decision}: RAG \linebreak \textit{Execution}: [1] ... [2] ... [3] ... The director of film Lettre Ouverte is not specified. Final answer: "Not specified" \linebreak \textit{Verification}: No & \textit{Decision}: Direct \linebreak \textit{Execution}: "Final answer: "Paris" \\ \hline
        Which film has the director who died later, Stolen Identity or Them!? & \textit{Decision}: Plan \linebreak \textit{Execution}: Follow up:  Who is the director of Stolen Identity? Intermediate Answer: Andrzej Wajda. Follow up: When did Andrzej Wajda die? Intermediate Answer: September 9, 2016. Follow up: Who is the director of Them!? Intermediate Answer: Gordon Douglas. Follow up: When did Gordon Douglas die? Intermediate Answer: September 30, 1993. Final answer: "Stolen Identity" \linebreak \textit{Verification}: No & \textit{Decision}: Direct \linebreak \textit{Execution}: "Final answer: "Them!" \\
        \bottomrule
    \end{tabular}
    \caption{Qualitative analysis depicting cases where \modelVerifyName{} improves over the first round outputs.}
    \label{tab:dyplan-verify-improvements}
\end{table*}

\subsection{Qualitative Examples for \modelVerifyName}
\label{sec:dyplan-verify-qualitative-examples}

In Table~\ref{tab:dyplan-verify-improvements}, we show some examples wherein additional verification identifies and rectifies any potential mistakes made in the first round.
Specifically, we also show examples (last two) wherein the first-round strategy was better but provided the wrong answer, while the second-round strategy was weaker but provided the right answer.
Overall, this shows how verification, along with multi-strategy selection, is a key component to improve model performance.

\begin{table*}[t]
    \centering
    \small
    \begin{tabular}{p{3cm}|p{5.6cm}|p{6cm}}
        \toprule
        \textbf{Question} & \textbf{Strategy 1 Output} & \textbf{Strategy 2 Output} \\
        \midrule
        \multicolumn{3}{c}{\textbf{Strategy 1: Direct \quad \quad Strategy 2: Reason \quad \quad Strategy 1 > Strategy 2}} \\
        \midrule
        Who died first, George Archainbaud or Ralph Murphy? & Final answer: "George Archainbaud" & George Archainbaud was born on October 15, 1899. Ralph Murphy was born on April 16, 1891. Therefore, Ralph Murphy died first. Final answer: "Ralph Murphy" \\ \hline
        Alice David is the voice of Lara Croft in a video game developed by which company? & Final answer: "Crystal Dynamics" & The video game developed by Eidos Interactive is Tomb Raider. The voice of Lara Croft in Tomb Raider is Camilla Luddington. Alice David is not the voice of Lara Croft. Final answer: "Eidos Interactive" \\
        \midrule
        \multicolumn{3}{c}{\textbf{Strategy 1: Plan \quad \quad Strategy 2: Reason \quad \quad Strategy 1 > Strategy 2}} \\
        \midrule
        When was the  former Indian cricketer who made his ODI debuts during the 2000 ICC KnockOut Trophy born? & Follow up:  Who is the former Indian cricketer who made his ODI debuts during the 2000 ICC KnockOut Trophy? Intermediate Answer: Virender Sehwag. Follow up: When was Virender Sehwag born? Intermediate Answer: Virender Sehwag was born on October 30, 1978. Final answer: "October 30, 1978". & The former Indian cricketer who made his ODI debuts during the 2000 ICC KnockOut Trophy is Harbhajan Singh. Harbhajan Singh was born on July 3, 1980. Final answer: "July 3, 1980" \\ \hline
        What is the birthplace of the Senator who represents the first of 62 districts in the State Senate? & Follow up: Who represents the first of 62 districts in the State Senate? Intermediate answer: Chuck Schumer. Follow up: Where is Chuck Schumer from? Intermediate answer: Chuck Schumer is from Brooklyn, New York. Final answer: "Brooklyn, New York" & The State of California has 40 Senate districts. The first of these is the 1st Senate district. The Senator representing the 1st Senate district is Toni Atkins. Toni Atkins was born in San Diego, California. Final answer: "San Diego, California" \\ 
        \midrule
        \multicolumn{3}{c}{\textbf{Strategy 1: Reason \quad \quad Strategy 2: Retrieval \quad \quad Strategy 1 > Strategy 2}} \\
        \midrule
        Are Random House Tower and 888 7th Avenue both used for real estate? & Random House Tower is a residential building. 888 7th Avenue is the address of the General Motors Building, which is a commercial building. Final answer: "no" & [1] ... [2] ... [3] ... Random House Tower is used for office space. 888 7th Avenue is used for office space. Final answer: "yes" \\ \hline
        What is the name of the oldest child that attended the Trump campaign-Russian meeting? & The Trump campaign-Russian meeting refers to the meeting between Donald Trump Jr. and a group of Russians in June 2016. The oldest child who attended the meeting was Donald Trump Jr.'s son, Donald Trump III. Final answer: "Donald Trump III" & [1] ... [2] ... [3] ... The meeting was held on June 9, 2016. The meeting was attended by Donald Trump Jr., Jared Kushner, Paul Manafort, Natalia Veselnitskaya, and Rinat Akhmetshin. The oldest child that attended the meeting is Jared Kushner. Final answer: "Jared Kushner" \\ \hline
        Luke Null is an actor who was on the program that premiered its 43rd season on which date? & Luke Null is an actor who was on the program Saturday Night Live. Saturday Night Live premiered its 43rd season on September 30, 2017. Final answer: "September 30, 2017" & [1] ... [2] ... [3] ... Luke Null is an actor who was on the program "The Daily Show with Trevor Noah". The Daily Show with Trevor Noah premiered its 43rd season on September 28, 2021. Final answer: "September 28, 2021" \\
        \bottomrule
    \end{tabular}
    \caption{Qualitative cases eliciting odd behaviors when an inferior strategy yields the correct answer but a superior one fails highlighting the strategy hierarchy violations.}
    \label{tab:strategy-violation-examples}
\end{table*}

\subsection{Qualitative Examples for Strategy Hierarchy Violations}
\label{sec:appendix-strategy-hierarchy-examples}

In \S~\ref{sec:appendix-hierarchy-violations}, we discussed how violations in the hierarchy could contribute to significant model performance.
Here, we provide some qualitative examples for violations of various strategy combination sets in Table~\ref{tab:strategy-violation-examples}.
In the first comparison of Direct with Reason, we observe how reasoning leads to hallucinations or wrong logical inferences leading to the wrong final answer; while the model can answer correctly when prompted directly.
In the second comparison of Plan with Reason, we observe how breaking the questions into atomic questions helps the model to correctly answer for Planning. It's expected that the model can reason in a similar way, but in its reasoning, it again starts to hallucinate.
Finally, we show the case of Reason with Retrieval, wherein the model correctly answers the questions using its self-knowledge, but in the context of retrieved passages, the model suddenly starts to hallucinate or state incorrect facts.
On further analysis, we find that some of these cases can be attributed to wrong retrievals. However, many of them are just incorrect reasoning itself - which is quite odd and strange.
Overall, our work doesn't focus deeply on why such violations happen (which can be an area of future study), but we majorly provide the verification loop in \modelVerifyName{} to navigate through such failure cases.

\end{document}